%% file: main.tex
\documentclass[lettersize,journal]{IEEEtran}
\usepackage{amsmath,amsfonts}
\usepackage{array}
\usepackage{textcomp}
\usepackage{stfloats}
\usepackage{url}
\usepackage{verbatim}
\usepackage{graphicx}
\usepackage{cite}
\usepackage{algpseudocode}
\usepackage{amsmath}
\usepackage[nameinlink,capitalize]{cleveref}
\crefname{algocf}{Algorithm}{Algorithms}    % algorithm2e conflicts with cleveref
\usepackage{subfigure}  % subfigure and subfig are mutually exclusive
\usepackage{booktabs}
\usepackage[linesnumbered,ruled]{algorithm2e}   % algorithm2e and algorithm are mutually exclusive
\usepackage{xcolor}

\ifodd 1
\newcommand{\com}[1]{\textbf{\color{red}(COMMENT: #1)}} %comment of the 
\else

\newcommand{\com}[1]{}
\fi

\def\fig{Fig.}

\def\ie{i.e.}

\def\eq{Eq.}

\begin{document}

% \title{Boosting Collaborative Mobile Robot Olfactory Sensing with Resilience in Patchy Plumes}
% \title{GasHunter: Towards Practical Gas Source Localization by Collaborative Robots}
\title{SniffySquad: Patchiness-Aware Gas Source Localization with Multi-Robot Collaboration}

\author{Yuhan Cheng$^*$,
        Xuecheng Chen$^*$,
        Yixuan Yang,
        Haoyang Wang,
        Jingao Xu,
        Chaopeng Hong,\\
        Xiao-Ping Zhang,
        Yunhao Liu,
        Xinlei Chen
        % IEEE Publication Technology,~\IEEEmembership{Staff,~IEEE,}
        % <-this % stops a space
\thanks{Manuscript submitted November 2024. \textit{(Corresponding author: Xinlei Chen.)}}

\thanks{$^*$ indicates equal contribution.}        
    
\thanks{Yuhan Cheng, Xuecheng Chen, and Haoyang Wang are with Shenzhen International Graduate School, Tsinghua University, Shenzhen, China (email:  \{cyh22, chenxc21, haoyang-22\}@mails.tsinghua.edu.cn).}
\thanks{Yixuan Yang is with Electrical and Computer Engineering, Duke University, Durham, North Carolina, United States (email: yixuan.yang@duke.edu).}
\thanks{Jingao Xu is with School of Software, Tsinghua University, Beijing, China (email: xujingao13@gmail.com).}
\thanks{Chaopeng Hong is with Tsinghua Shenzhen International Graduate School, Tsinghua University, Shenzhen, China (email: hongcp@sz.tsinghua.edu.cn).}
\thanks{Xiao-Ping Zhang is with Shenzhen International Graduate School, Tsinghua University, Shenzhen, China, and also with RIOS Lab, Shenzhen, China (e-mail: xpzhang@ieee.org).}
\thanks{Yunhao Liu is with Global Innovation Exchange, Tsinghua University, Beijing, China, and also with the Department of Automation, Tsinghua University, Beijing, China (email: yunhao@tsinghua.edu.cn).}
\thanks{Xinlei Chen is with Shenzhen International Graduate School, Tsinghua University, Shenzhen, China, and also with Pengcheng Laboratory, Shenzhen, Guangdong China, and RIOS Lab, Shenzhen, Guangdong, China (e-mail: chen.xinlei@sz.tsinghua.edu.cn).}
}


% % column for its text to clear the IEEEpubid mark.

\maketitle

\begin{abstract}
Gas source localization is pivotal for the rapid mitigation of gas leakage disasters, where mobile robots emerge as a promising solution. However, existing methods predominantly schedule robots' movements based on reactive stimuli or simplified gas plume models. These approaches typically excel in idealized, simulated environments but fall short in real-world gas environments characterized by their patchy distribution. In this work, we introduce \textit{SniffySquad}, a multi-robot olfaction-based system designed to address the inherent patchiness in gas source localization. \textit{SniffySquad} incorporates a patchiness-aware active sensing approach that enhances the quality of data collection and estimation. Moreover, it features an innovative collaborative role adaptation strategy to boost the efficiency of source-seeking endeavors. Extensive evaluations demonstrate that our system achieves an increase in the success rate by $20\%+$ and an improvement in path efficiency by $30\%+$, outperforming state-of-the-art gas source localization solutions.

\end{abstract}


\input{introduction-20241011}

\input{background}
\input{formulation}
\input{systemmodel}
\input{algorithm}

\input{evaluation}
\input{conclusion}

% \section*{Acknowledgments}
% This paper was supported by the National Key R&D program of China No. 2022YFC3300703, the Natural Science Foundation of China under Grant No. 62371269. Guangdong Innovative and Entrepreneurial Research Team Program No. 2021ZT09L197, Shenzhen 2022 Stabilization Support Program No. WDZC20220811103500001, and Tsinghua Shenzhen International Graduate School Cross-disciplinary Research and Innovation Fund Research Plan No. JC20220011.

% {\appendix[Proof of the Zonklar Equations]
% Use $\backslash${\tt{appendix}} if you have a single appendix:
% Do not use $\backslash${\tt{section}} anymore after $\backslash${\tt{appendix}}, only $\backslash${\tt{section*}}.
% If you have multiple appendixes use $\backslash${\tt{appendices}} then use $\backslash${\tt{section}} to start each appendix.
% You must declare a $\backslash${\tt{section}} before using any $\backslash${\tt{subsection}} or using $\backslash${\tt{label}} ($\backslash${\tt{appendices}} by itself
%  starts a section numbered zero.)}



\bibliographystyle{IEEEtran}
\bibliography{main}

\vfill

\end{document}

%% file: introduction-20241011.tex
\section{Introduction}

% \IEEEPARstart{L}{ocalizing} the gas leak source rapidly is of paramount importance. 
% Unforeseen gas leaks can swiftly create a highly flammable atmosphere, which will lead to risks of explosions, health hazards, as well as long-term environmental impacts, especially when the gas is toxic \cite{liu2017individualized, pal2018vehicle, maag2018w}. For instance, the Bhopal gas tragedy exposed over 500,000 people to methyl isocyanate (MIC) \cite{mishra2009bhopal}, while the Aliso Canyon blowout released over 100,000 tonnes of greenhouse gas methane into the atmosphere \cite{conley2016methane}. 
% It is crucial to note that only after locating the source can subsequent mitigation measures, such as shutting off valves and effectively containing or sealing the leaks, be executed \cite{liu2017delay}.

Rapid and accurate responses to gas leak incidents are essential for safeguarding human and environmental health, as leaked gases can rapidly create highly flammable or toxic conditions, posing significant risks of explosions and poisoning~\cite{liu2017individualized, pal2018vehicle, maag2018w}. For instance, in the United States alone, 2,600 gas leakage incidents have been reported, with 328 resulting in explosions and 122 fatalities~\cite{sun2024gastag}. A key aspect of quick response requires localizing the gas source, which involves analyzing the concentration and distribution of the gas in the air to trace it back to its origin. With the knowledge of source locations, subsequent mitigation operations, such as shutting off valves or sealing the leaks, can be conducted more logically, efficiently, and safely~\cite{liu2017delay}.

% Nonetheless, gas source localization (GSL) in emergent scenarios, like those mentioned above, has not been effectively resolved \cite{motlagh2023unmanned, abughanam2023localization}. 
% Specifically, extensive and complex pipelines offer numerous potential points of leakage, rendering it impractical to install sensors at every juncture \cite{ojeda2020evaluation}. 
% Furthermore, assigning human operators to engage in these laborious and perilous activities not only increases the risk of misinterpretation \cite{arain2021sniffing}, but also poses a risk to their health and lives \cite{ojeda2023robotic}.

Conventional gas source localization (GSL) solutions fall into two categories: 
($i$) human expert-based solutions assign human operators to engage in affected areas. These laborious and perilous activities not only increase the risk of misinterpretation but also imperil the safety of the operators~\cite{mccarron2018air};
and ($ii$) wireless sensor network (WSN)-based methods utilize preinstalled static sensors to detect the gas source by monitoring gas concentration readings~\cite{manes2016realtime}. However, this approach is constrained by spatial resolution limitations, particularly in environments with extensive and intricate pipeline networks that present numerous potential leakage points.

% Deploying autonomous mobile olfactory robots for GSL presents a solution with great potential \cite{iyer2019living, moradi2018skycore, khochare2021heuristic}, as they can search the environment with high efficiency without risking human lives \cite{xu2022swarmmap, zhou2022swarm}. 
% We take gas leakage in a factory with intricate pipelines as an example as shown in \cref{fig:scenario}. Here, a fleet of robots equipped with gas concentration sensors and wind sensors are dispatched to search the surroundings and gather gas sensory data. By analyzing and interpreting the environmental conditions, the robots autonomously navigate to pinpoint the source of the gas leak.

% Fortunately, deploying multiple mobile olfactory robots presents a promising solution, as the robots can search the environment with high efficiency without risking human lives. We take gas leakage scenario in a factory with intricate pipelines as an example, as shown in \cref{fig:scenario}. Here, a fleet of robots equipped with gas concentration sensors and wind sensors are dispatched to search the surroundings and gather gas sensory data. By analyzing and interpreting the environmental conditions, the robots autonomously navigate to pinpoint the source of the gas leakage.

In this work, we aim to devise an effective strategy for collaboratively scheduling multiple olfactory robots to localize the gas source in real-world environments. 
Particularly, we utilize multiple robot's activeness and collaborative capability to gather more information efficiently, enhancing the synergy between their sensing and scheduling abilities. 
% \rev{Particularly, we fully utilize multiple robot's activeness and collaborative capability to gather more information, thereby enhancing accurate and efficient GSL across complex environments.}

\begin{figure}[t]
    \setlength{\abovecaptionskip}{0.0cm}
    \setlength{\belowcaptionskip}{-0.4cm}
    \centering
    \includegraphics[width=0.9\columnwidth]{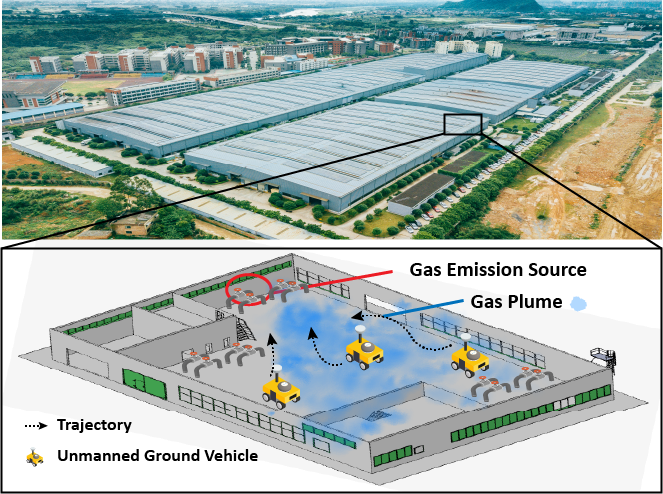}
    
    \caption{Mobile olfactory robots autonomously search for and navigate towards the source of gas leakage. }
    \label{fig:scenario}
\end{figure}

We take gas leakage in a factory with intricate pipelines as an example scenario, as shown in \cref{fig:scenario}. Here, a fleet of robots equipped with gas concentration sensors and wind sensors are dispatched to search the surroundings and gather gas sensory data. By analyzing and interpreting the environmental conditions, the robots autonomously navigate to pinpoint the source of the gas leakage. However, translating this idea into a practical system is non-trivial and faces two challenges:

% \noindent $\bullet$ \textbf{The patchy nature of gas plumes confuses and traps the robots.}
% In reality, the gas landscape is characterized by a \textit{patchy} structure. As demonstrated by field measurements of gas concentration (see \cref{fig:conc_distr_characteristic}), gas plumes in realistic atmospheric environments fragment into disjointed patches and form distinct and scattered areas. 
% From the perspective of source-seeking robots, the concentration measured along the horizontal centerline exhibits pronounced fluctuations as the distance to the source increases, since gas patches are separated by regions where gas concentration is below detectable levels. 
% This intermittent characteristic introduces uncertainty in the source direction when robots devise paths using noisy and local sensory data, ultimately leading to their entrapment within these patches and failure to complete GSL.
% Existing solutions, including bio-inspired \cite{reddy2022olfactory, chen2019odor} and probabilistic model-based approaches~\cite{rhodes2023autonomous, rhodes2020informative}, either rely on local gas concentration gradient or oversimplified gas dispersion models, limiting their effectiveness to ideal scenarios with continuous concentration gas fields.

\begin{figure}[!t]
    \setlength{\abovecaptionskip}{0.0cm}
    \setlength{\belowcaptionskip}{-0.1cm}
    \centering
    \includegraphics[width=1.0\columnwidth]{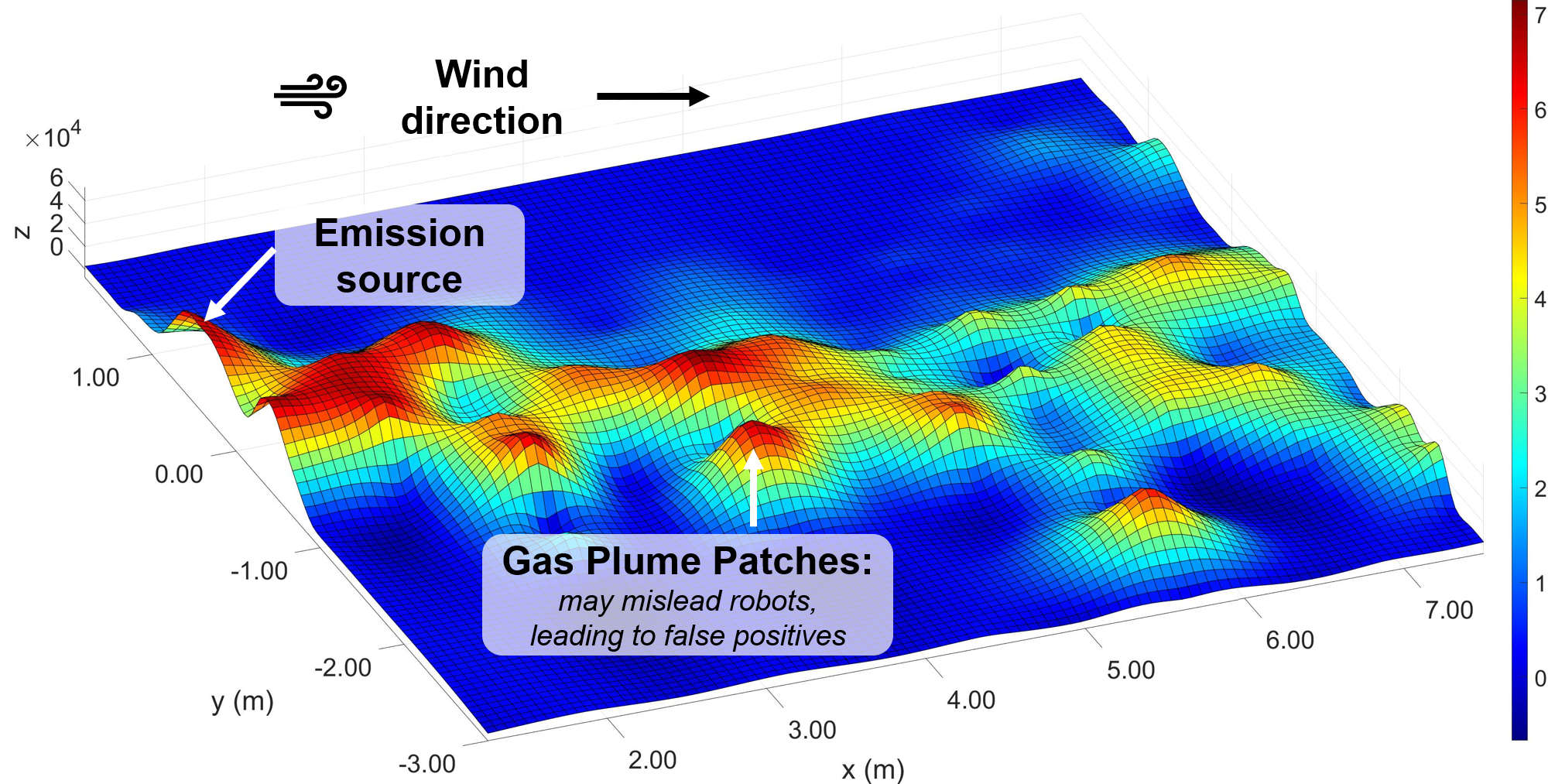}
    % \caption{Real-world spatial characteristics of gas concentration distribution are observed within a sub-area of our testbed. A gas emission device is positioned on the left side, and winds blow from left to right. The heatmap illustrates concentration values, representing the number of particles with a diameter $>$0.3um in 0.1L of air. The gas plume patches may mislead source seeking robots into incorrectly identifying these (false positives) as the actual source of the gas emission. }
    \caption{\textbf{Spatial characteristics of gas concentration}. We conducted a proof-of-concept experiment to check gas characteristics in our indoor testbed, with a gas emission device at $(x,y)=(1.0,0.0)$ and winds blowing along the $x$ axis. The heatmap illustrates concentration values, representing the number of particles with a diameter $>$0.3um in 0.1L of air. The observed gas plume patches may mislead source-seeking robots into falsely identifying them as the actual gas emission source.} 
    \label{fig:conc_distr_characteristic}
\end{figure}

\noindent $\bullet$ \textbf{The patchy nature of gas plumes confuses and traps the robots.}
The gas landscape is characterized by a \textit{patchy structure}~\cite{bourne2020decentralized, hutchinson2018entrotaxis}, as evidenced by field measurements of gas concentration (see \cref{fig:conc_distr_characteristic}). As seen, gas plumes fragment into disjointed patches and form distinct and scattered areas. 
From the perspective of source-seeking robots, the concentration measured along the horizontal centerline exhibits pronounced fluctuations as the distance to the source increases, since gas patches are separated by regions where gas concentration is below detectable levels. 
This intermittent characteristic introduces uncertainty in determining the source direction when robots devise paths using noisy and local sensory data, ultimately leading to their entrapment within these patches and failure to complete GSL.
Existing solutions, including bio-inspired \cite{reddy2022olfactory, chen2019odor} and probabilistic model-based approaches~\cite{rhodes2023autonomous, rhodes2020informative}, either rely on local gas concentration gradient or oversimplified gas dispersion models, rendering them effective only in idealized scenarios with continuous gas concentration fields.

\noindent $\bullet$ \textbf{Trade-off between source localization effectiveness and search efficiency for collaborative robots.}
Since gas distribution varies continuously over space and time, it is impossible to measure at every possible location all the time. Therefore, robots must strike a balance between two conflicting objectives within limited sampling: ($i$) identifying new potential gas source positions and ($ii$) excluding false positive source positions. Achieving these goals simultaneously complicates the localization of the true gas source efficiently and effectively. 
Specifically, the former goal necessitates extensive exploration and sampling of gas data, while the latter goal requires utilizing previously acquired information to reach and verify the estimated source position's probability.
Current multi-robot GSL solutions tend to exhibit either clustered or dispersed collaborative behaviors, thereby undermining the simultaneous achievement of both objectives~\cite{bourne2019CoordinatedBayesianBasedBioinspired, chen2019odor, duisterhof2021sniffy}.

\noindent \textbf{Remark}. As far as we are aware, previous works develop their operating principles ignoring the patchy nature of gas plumes, based on which they schedule agents without fully harnessing their versatility and collaborative capabilities at a system level.

To tackle the above challenge, we design and implement \textit{SniffySquad}, a collaborative olfactory robot scheduling system for gas source localization. Benefiting from \textit{SniffySquad}, a team of olfactory robots can adapt to field patchiness and dynamically adjust their movements accordingly, enabling efficient and effective emission source localization. In general, \textit{SniffySquad} excels in the following two aspects.

% % To achieve individual-level resiliency, we propose a \textit{Patchiness-aware Active Sensing} method. It improves the collected data effectiveness (the quality of sampled data) by movement coordination. This is achieved by adjusting the robots' gradient-based moving direction with a regulation term. This term injects the motion with an appropriate amount of perturbation, enabling them to escape from false positive source positions efficiently.
% \noindent $\bullet$ At the individual level, we propose a \textit{Patchiness-aware Active Sensing} method, which builds on probabilistic source estimation and embeds principles of Langevin MCMC, to improve the quality of collected data and estimation. % (integrates deterministic and stochastic components of sensing direction)
% This is achieved by incorporating a regulatory term into the robots' gradient-based sensing direction. This enables escape from false positive source positions, resulting in the collection of more valid and informative data.

\noindent $\bullet$ \textbf{At the individual level}, we propose a \textit{Patchiness-aware active sensing} method, which refines the probabilistic source estimation by incorporating the patchy characteristic of gas in each robot's moving strategy. Inspired by the principles of Langevin MCMC \cite{deng2020contourSGLD}, we adjust robots' gradient-based moving direction with a regulation term, allowing them to escape false positive source positions and thereby collect more informative sensing data.

% % To implement a system with team-level resiliency, we devise a \textit{Potential-instructed Collaborative Roles Adaptation} strategy for multiple robots. It enhances source-seeking efficiency and accelerate the GSL process. 
% \noindent $\bullet$ At the team level, we devise a \textit{Potential-instructed Collaborative Roles Adaptation} strategy, which leverages both external environmental cues and internal team dynamics, to enhance source-seeking efficiency and accelerate the GSL process. 
% Specifically, robots in the team collaboratively estimate the potential of each position locating the emission source. 
% Those positioned in areas with higher probabilities conduct fine-grained search to inspect spurious concentration maxima (exploiter), while other robots perform coarse-grained search to discover new potential sources (explorer). 
% When an explorer finds potential sources with higher probability than an exploiter, they exchange their roles to allow flexible collaboration. %, so that the explorer becomes an exploiter while the exploiter is released for course-grained search. 
% % In this manner, the robots' roles are adjusted adaptively based on their current positions and the potential to localize the source in the surroundings, thus parallelizing the exploration of the environment and exploitation of collected information in a flexible manner.

\noindent $\bullet$ \textbf{At the team level}, we design a \textit{Potential-instructed collaborative roles adaptation} strategy that further enhances the source-seeking efficiency by adjusting robots' roles 
based on their spatial distribution and measurements in the past. 
% based on their current spatial distribution and potential to localize the source. 
Each robot in the team adopts either a fine-grained search role to inspect spurious signals (\ie, exploiter) or a coarse-grained search role to discover potential new sources (\ie, explorer). These roles are adaptively adjusted based on the estimated probabilities of each robot's proximity to the emission source, thereby parallelizing the exploration of the environment and exploitation of collected information in a flexible manner.

% % To the best of our knowledge, we are the first to handle the turbulent airflow and patchy gas plume issue from the view of nonconvex learning. 
% We validate our design with experiments on a multi-robot testbed in a $15m \times 10m$ space and large scale physical feature based simulations in two different scenarios. Experiment results show that our system achieves a $20\%+$ success rate improvement and a $30\%+$ path efficiency improvement compared to state-of-the-art methods~\cite{reddy2022olfactory,ojeda2021information,vergassola2007infotaxis}.
% % We implemented the proposed system on an indoor gas emission testbed, in a  $15m \times 10m$ space. We build a team of olfactory unmanned ground vehicles to conduct the gas source localization task. 
% % We further conduct comprehensive simulation experiments with various number of robots in two different scenarios (an empty space and a structured environment). 
% % We compare our system with the state-of-the-art (SOTA) of both bio-inspired and probabilistic GSL algorithms, Surge-Cast \cite{reddy2022olfactory} and Infotaxis \cite{ojeda2021information,vergassola2007infotaxis}. The experiment results show that our system achieves success rates higher than baselines by $> 20\%$, outperforming then in path efficiency by $> 32.84\%$. 
% % We then examine the robustness of our system under various conditions and finally conduct an ablations study to understand each module in our system. 

We evaluate the performance of \textit{SniffySquad} and compare it with state-of-the-art baseline methods through experiments on a real-time multi-robot testbed (12 hours) and extensive physical feature-based simulations (750 runs). Experiment results show that our system outperforms all baselines, achieving a $20\%+$ success rate improvement and improving path efficiency by $>30\%$.

We summarize the contributions of this paper as follows:
% \begin{itemize}
% \item We design \textit{SniffySquad}, a collaborative multi-robot olfactory sensing system for gas source localization, specifically designed to handle the patchy characteristic in realistic gas plume environments.
% \item We devise a patchiness-aware active sensing method, which builds on probability source estimation and embeds principles of Langevin dynamics, to improve the quality of collected data and estimation.
% \item We develop a potential-instructed collaborative roles adaptation strategy, which leverages both robots' spatial distribution and external environmental cues, to enhance source-seeking efficiency and accelerate the GSL process. 
% \item Extensive experiments are conducted for validation and evaluation in both a real-world testbed and a physical-feature-based gas dispersion simulator. 
% \end{itemize}

\begin{itemize}
\item We propose \textit{SniffySquad}, a collaborative multi-robot olfactory sensing system for accurate and efficient gas source localization, specifically designed to handle the patchy characteristics in real-world gas fields.
\item We introduce a patchiness-aware active sensing method that refines probabilistic source
estimation by incorporating gas patchiness in each robot’s moving strategy. Building on this design, we further devise a potential-instructed collaborative roles adaptation strategy, which parallelizes the exploration of the environment and exploitation of collected information in an adaptive manner.
\item We develop a prototype system and evaluate \textit{SniffySquad} through a real-world testbed and a physical-feature-based gas dispersion simulator. Extensive evaluation results show its effectiveness and superior performance. 
\end{itemize}

The remainder of the paper is organized as follows. We first introduce the background, motivation, and related works in \cref{sec:background}. In \cref{sec:preliminaries}, we present preliminaries, followed by the system overview in \cref{sec:systemmodel}. In \cref{sec:algo1,sec:algo2}, we introduce the algorithm design, involving detailed descriptions of the patchiness-aware active sensing and potential-instructed collaborative roles adaptation. \cref{sec:eval} showcases the implementation and evaluation. Finally, \cref{sec:conclusion} concludes the paper.

% \begin{figure}[!t]
%     \setlength{\abovecaptionskip}{0.2cm}
%     \setlength{\belowcaptionskip}{-0.1cm}
%     \centering
%     \includegraphics[width=1.0\columnwidth]{figs/fig2.png}
%     \caption{Spatial characteristics of the gas concentration distribution.}
%     \label{fig:conc_distr_characteristic}
% \end{figure}

%% file: background.tex
\section{Background and Related Work} \label{sec:background}

\subsection{Characteristics of Gas Concentration Distribution}
% \begin{figure*}[t]
%     \centering
%     % \includegraphics[width=0.4\columnwidth]{figs/Qheatmap.png}
%     % \includegraphics[width=0.4\columnwidth]{figs/conc_over_space.png}
%     % \includegraphics[width=1.0\columnwidth]{figs/conc_over_time.jpg}
%     \includegraphics[width=2.1\columnwidth]{figs/gas_concentration_over_time_final.png}
%     \caption{Spatial and temporal Characteristics of the Gas concentration distribution.}
%     \label{fig:conc_distr_characteristic}
% \end{figure*}
Spatial characteristics of gas concentration in real-world are depicted in \cref{fig:conc_distr_characteristic}. The data are measured in our testbed, which will be introduced in \cref{subsec:experiment_setting}. \cref{fig:conc_distr_characteristic} shows that gas distribution exhibits patchiness, which is caused by turbulent atmosphere stretching and fragmenting plumes into disjointed patches spatially \cite{bourne2020decentralized}. This characteristic may mislead source-seeking robots into falsely identifying them as the actual gas emission source.

Note that characteristics of gas concentration distribution described by models commonly used in previous gas source localization (GSL) works differ significantly from those observed in practical environments. We first introduce the model that existing work relies on. 
To model the concentration of a substance flowing in a fluid, the transportation process is typically adopted, which is described by the partial differential convection-diffusion equation $\frac{\partial c}{\partial t}=\nabla \cdot \mathbf{u} c -\nabla \cdot \Gamma \nabla c + S$. Here scalar $c$ is the gas concentration, $\mathbf{u}$ is the velocity vector $(u,v,w)$, $\nabla$ and $\nabla \cdot$ represents the gradient and the divergence of the $c$, and $S$ is the source of a scalar. 
While this equation can be employed for numerical simulation in computational fluid dynamics (CFD), as highlighted in the preceding section, probabilistic gas source localization approaches necessitate an analytical model rather than a transportation process for numerical simulation. To this end, previous works \cite{bourne2019CoordinatedBayesianBasedBioinspired,loisy2022searching} solve a simplified transportation equation by assuming steady-state conditions and assuming no degradation, yielding the Gaussian plume model \cite{holzbecher20122d}. 

% However, this oversimplified model can only be used to model gas concentration landscapes at microscopic scales, where molecular diffusion generate smooth changes in concentrations of chemicals \cite{reddy2022olfactory}. At distances on the order of meters or larger, gas plumes are significantly dispersed, resulting in patchy and discontinuous plumes as macroscopic scales. 
% A realistic gas concentration distribution is demonstrated in \cref{fig:conc_distr_characteristic}. 
However, this oversimplified model can only be used to model gas concentration landscapes at microscopic scales, where molecular diffusion generates smooth changes in concentrations of chemicals \cite{reddy2022olfactory}. In contrast, as shown in \cref{fig:conc_distr_characteristic}, at distances on the order of meters or larger, gas plumes are significantly dispersed, resulting in patchy and discontinuous plumes as macroscopic scales. 
% Besides, the temporal attribute of the gas concentration landscape is time-variant and fluctuated, as the gas plume is carried by the wind flow. The concentration at a fixed position varies over time, as shown in \cref{fig:conc_distr_characteristic} (c). This temporal characteristic of the concentration field aligns with the patchiness of the plume, as the patches moves in and out a particular position.  

% \subsection{Related Work} \label{subsec:related_work}
\subsection{Mobile Robot Olfaction-based Gas Source Localization}
Current approaches can be categorized into two classes, which will be introduced first, reactive bio-inspired and probabilistic algorithms. Then multi-robot gas source localization methods are summarized, introducing how collaboration is incorporated. 
% Previous works either do not explicitly consider the airflow turbulence (e.g. bio-inspired methods \cite{reddy2022olfactory}), or rely on oversimplified hence unrealistic gas plume models (e.g. Gauss plume) \cite{vergassola2007infotaxis, bourne2019CoordinatedBayesianBasedBioinspired, bourne2020decentralized}.

\subsubsection{Reactive bio-inspired GSL}
Bio-inspired algorithms draw inspiration from the behavior of insects or rodents. They involve chemotaxis, anemotaxis, or combining both \cite{reddy2022olfactory}.
For chemotaxis, such strategies have the robot turn to the opposite direction if it experiences a declining odor concentration gradient along its trajectory. Chemotaxis fails when odor trails are broken, olfactory stimuli are sporadic, and consequently gradients are absent. 
Anemotaxis-driven navigation can be divided into surge (the robot moves upwind), casting, spiral, and zigzagging \cite{chen2019odor}. For casting, a robot in the plume moves upwind until it loses the plume, then it turns and moves cross-wind until it hits odor packets again. When a robot loses the plume, it moves along a spiral or zig-zag to reacquire the plume. 
Bio-inspired methods are reactive in nature, making them susceptible to intermittent gas concentration signals caused by spurious gas patches. 

\subsubsection{Probabilistic gas source localization methods}
% Secondly, probabilistic algorithms estimate the source location utilizing the Bayesian filter. While these methods are designed to mitigate the sensitivity to unstable atmospheric conditions or other natural disturbances, they often rely on oversimplified hence unrealistic gas plume models (e.g. Gauss plume) \cite{vergassola2007infotaxis, bourne2019CoordinatedBayesianBasedBioinspired, bourne2020decentralized}. Although \cite{ojeda2023robotic} adopts a relatively practical dispersion model, it relies on predictions of unvisited locations' measurements, which are inaccurate. 

% Parametric/Non-parametric mathods
According to models used for the environment's representation, probabilistic gas source localization methods can be categorized into parametric and non-parametric algorithms \cite{rhodes2023autonomous}. 

\noindent $\bullet$ \textbf{Parametric algorithms}, which rely on an atmospheric transport and diffusion (ATD) model mapping each point in the environment to the expected gas concentration there, assume a finite set of parameters. In the context of GSL, the set of parameters is called source term, involving a combination of the position of the source, the release rate, the downwind direction and velocity, and 3-dimensional diffusion terms in different directions (x, y, and z). Prevalent plume models include the isotropic plume model \cite{vergassola2007infotaxis} and the Gaussian plume model \cite{wang2018efficient}. 

\noindent $\bullet$ \textbf{Non-parametric algorithms}. They model the spatial distribution of the gas plume rather than focus on terms parameterizing the gas source, thus can not be defined in terms of a finite set of parameters. 
These algorithms originate from the problem of gas distribution mapping (GDM). Popular algorithms include Kernel DM+V \cite{lilienthal2009statistical} (and its variants) and Gaussian Markov random fields (GMRF) based spatial modeling \cite{gongora2020joint,rhodes2020informative}. 

% Pros & Cons of Parametric/Non-parametric mathods
Parametric methods are mostly computationally efficient and exhibit simple analytical forms. The simplicity of the plume model, on the other hand, ignores the stochasticity of gas emissions and the spatial nonuniformity of wind fields. It renders this type of algorithms idealistic. Moreover, these methods fail under the existence of multiple sources or the inclusion of walls and obstacles. The latter fact restrains the methods' applicability to indoor places, such as factories. 
In contrast, non-parametric algorithms make less assumptions about the atmospheric environment and the gas source characteristics, hence are much more flexible to be applied in various scenerios. However, without much prior knowledge embedded in non-parametric algorithms, inference at areas that are previously unvisited are typically inaccurate. Moreover, the extensive parameter space poses computational challenges. 

% gradient/information-based
According to planning strategies used to seek the source, probabilistic GSL methods can be categorized into gradient-based and information-based algorithms. 

\noindent $\bullet$ \textbf{Gradient-based probabilistic GSL planners} resemble the idea of chemotaxis but differ from it in the way they compute gradients. 
In chemotaxis, robots utilize merely measurements along its trajectory to identify whether the concentration is rising or declining. In probabilistic GSL algorithms, robots can leverage the probability model to calculate the direction, either analytically or numerically. 
For example in \cite{bourne2019coordinated}, particle filter biased random walk (BRW) modifies BRW \cite{marques2002olfaction} to utilize an analytical gradient computed from source term estimation results based on the Gaussian plume model. 

\noindent $\bullet$ \textbf{Information-based GSL planners} select actions to maximize the expected information gain (i.e., the uncertainty reduction). The algorithm can be implemented by optimizing upper confidence bound (UCB) \cite{cao2023catnipp}, entropy such as mutual information \cite{Ma2023GaussianME}, or Fisher information \cite{Zhang2022DistributedIS}. 

\subsection{Cooperative Gas Source Localization methods}
An apparent way of increasing GSL efficiency and accuracy is by utilizing multiple robots as a team. Some directly extend single robot gas source localization algorithms by spanning the action space. Other methods, tailored for cooperative GSL, are listed here. 
($i$) Particle Swarm Optimization (PSO)-based algorithms \cite{duisterhof2021sniffy} are inspired by the bird flock searching for food. If an agent discovers a better pattern (potential food source), others adjust their movements to follow it. When robot teams adopt this strategy for GSL, they will have a tendency to move towards the same local optimium and are considered to lack cooperation \cite{chen2019odor}. 
($ii$) Another type of approach \cite{bourne2019CoordinatedBayesianBasedBioinspired} achieves cooperation by dispatching robots to investigate several likely source positions separately, so that they can either be ruled out or considered more thoroughly. 

In summary, these multi-robot gas source localization methods induce either gathering or scattering behavior, so they lack efficiency in achieving both goals. Technically, each individual in the team acts similarly and are not assigned heterogeneous roles. 
In contrast, our work empowers team members to adopt different but complementary roles, enhancing individual diversity and overall synergy. This distinction highlights the advantage of a more strategic and versatile approach to team coordination.

%% file: formulation.tex
\section{Preliminaries} \label{sec:preliminaries}
\subsection{Definitions}
We denote gas source position and robot's position at time $t$ as $x_s$ and $x^{(i)}(t)$, respectively. The total time spent for gas source localization is represented by final time step $t_f$. The trajectory of a robot and its corresponding sensory measurements are denoted by $\mathbf{z}_{1:t}$ and $\mathbf{x}_{1:t}$. 

We express measurements at position $x$ as $z(x)$. We denote measurement data as $z=(z^c,z^w)$, where gas concentration and airflow direction are denoted by $c$ and $\mathbf{u}=(u,v,w)$, respectively. 
The system is only accessible to a belief $b(t)$ incorporating sensory measurements $\mathbf{z}_{1:t}$ along past trajectories $\mathbf{x}_{1:t}$ due to partial observability. 
It follows the Bayesian estimation framework by continually updating the likelihood of each position being situated within the gas plume region based on incoming measurements $z$ gathered from sensors equipped on the mobile robots. Mathematically, $p(x | z_{1:t}) \propto p(x | z_{1:t-1}) p(z_t | x,z_{1:t-1})$. 

Important symbols used in the paper are listed in \cref{tab:notation}. 

\subsection{Problem Formulation}
The goal of \textit{SniffySquad} is to minimize the search time, defined as the duration from departure to a robot reaching the vicinity of a gas source. A search trial terminates when a robot's position is within a predefined threshold of the gas source or the predefined time limit is reached. Mathematically, a successful search indicates that $\exists {i \in \{1, 2, \ldots, M\}} \ \text{s.t.}\ {\Vert x^{(i)}(t_f) - x_s \Vert} \leq d_\varepsilon$, where $d_\varepsilon$ is the distance threshold. Therefore, the system objective can be formulated as:
\begin{equation}
    \begin{aligned}\label{eq:formulation}
    \min_{\mathbf{x}(t), \mathbf{u}(t)} \quad & t_f & \\
    \mbox{s.t.}\quad
    & b(t) = bel(\mathbf{z}_{1:t}, \mathbf{x}_{1:t}),  & \\
    & u^{(i)} (t) = g\left(x^{(i)} (t), x^{(-i)} (t), b(t)\right),  & \\
    & \dot{x}^{(i)} = f(x^{(i)}, u^{(i)}),  & \\
    % & \tau^{(i)} = \mathfrak{T}(x^{(i)}, b, \tau)  & \\
    & \mathbf{x}(t_0)=\mathbf{x}_0, \mathbf{x}(t_f)=\mathbf{x}_f, & \\
    & x\in\mathcal{X}, u\in \mathcal{U}, t_f > 0, & \\
    \end{aligned}
\end{equation}

The global gas source estimator is represented as $bel(\cdot)$. The collaborative planner is denoted as $g(\cdot)$, which determines the movements of robot $i$, $u^{(i)}(t)$, according to the estimation results $b(t)$, its own position$x^{(i)}(t)$, and other robots' positions $x^{(-i)}(t)$. The control logic is denoted as $f(\cdot)$, describing the dynamics of specific robots. 

%% file: systemmodel.tex
% \section{System Model and Problem Statement} \label{sec:systemmodel}
\section{System Overview} \label{sec:systemmodel}

\begin{figure}[!t]
\centering
\includegraphics[width=1.0\columnwidth]{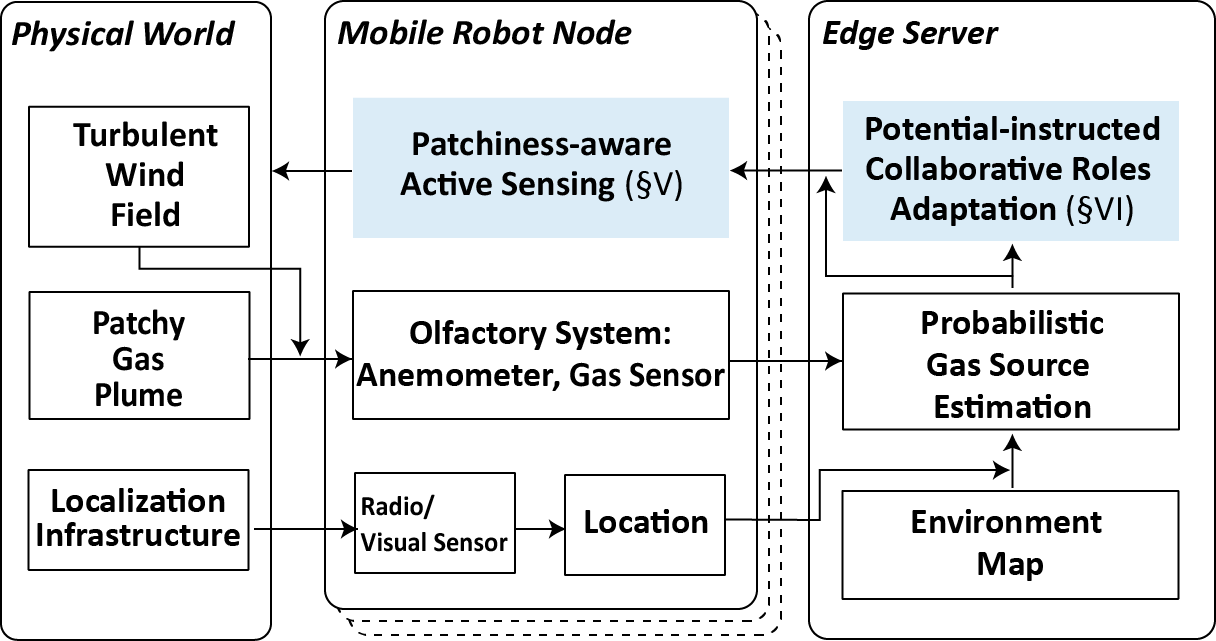}
\caption{System overview.}
\label{fig:systemmodel}
\end{figure}

In this part we describe the system model and interactions between various components. 
The design objective of our system is to schedule a team of autonomous mobile robots equipped with gas concentration sensors for gas source localization in atmospheric turbulence, achieving rapid and robust navigation towards the source of leakage. In this work, the system interacting with the underlying physical world is composed of $M$ mobile robot nodes and an edge server. An overview of the system is shown in \cref{fig:systemmodel}. 

\noindent $\bullet$ \textbf{On the Mobile layer}, the mobile robot node consists of task-specific sensors in the olfactory system, infrastructure for self-localization (e.g., radio or visual sensors), and a patchiness-aware active sensing module. 
First, it features task-specific sensors for the olfactory system, including a gas sensor to measure gas concentration and an anemometer to measure airflow direction. 
% In terms of sensing, there are two task-specific modalities of environmental data in gas source localization, i.e., gas concentration and airflow direction. To obtain sensory measurements, each mobile robot node is equipped with an anemometer and a gas sensor. 
Second, the localization infrastructure utilizes an attached UWB tag to determine the robot's position by measuring distances to pre-deployed UWB anchors. 
% Pre-deployed localization infrastructure is assumed available. Consequently, the robots are assumed to be aware of their location in the environment, utilizing an attached UWB tag to measure distances between UWB anchors. 
Finally, the \textit{patchiness-aware active sensing module} (\S\ref{sec:algo1}) takes high-level plans (coordination) from the edge server as input, then outputs low-level control commands executed by motors. 

\noindent $\bullet$ \textbf{On the Edge layer}, The edge server serves as the platform coordinating the robot team by maintaining a environmental map, estimating the gas source, and adapting team roles collaboratively. 
First, it maintains a global probability map associating each grid in the environment map with the probability of 'the gas source locates at that position'. The edge server fuses local measurements, including gas, wind, and location data, and continually updates the likelihood of each position being situated within the gas plume region based on incoming measurements. 
Second, the \textit{potential-instructed collaborative roles adaptation module} (\S\ref{sec:algo2}) coordinate all robots' behaviors and dispatches motion plans to mobile robot nodes. 

\textbf{Workflow.} To summarize, robots in the system first take sensory measurements as inputs, including the gas concentration, wind speed and direction. The onboard measurements are filtered to update the probabilistic gas source estimation. 
Thereafter, based on the up-to-date potential function and current positions of robots in the team, the robots' roles are reconfigured systematically if their potential of locating at the source mismatches with their roles. 
Finally, the robots choose their sensing directions according to the patchiness-aware active sensing module. 
Subsequent contents delve into details of these components.

\begin{table}[!t]
    \caption{List of Important Notations Used in the Paper}
    \label{tab:notation}
    \centering
    \begin{tabular}{cp{0.7\columnwidth}}
    % \begin{tabular}{cl}
        \toprule
        Notation & Description \\
        \midrule
         \( M \) & Total number of robots \\
         \( x(t) \in \mathcal{X} \) & Position coordinates at time $t$ \\
         \( \mathbf{x} \) & Positions of all $M$ robots $= \begin{bmatrix} x^{(1)}, x^{(2)}, \ldots, x^{(M)} \end{bmatrix}^T$ \\
         \(x_s\) & Gas source position \\
         \( c \) & Gas concentration \\
         % \( c, \mathcal{C} \) & Gas concentration, concentration map \\
         \( \mathbf{u} \) & Wind velocity vector $=(u,v,w)$ \\
         \( z \in \mathcal{Z} \) & Sensory measurement collected from mobile robot nodes (consisting of gas concentration $z^c$ and the wind vector $z^w$) $=(z^c,z^w)$ \\
         % \( \mathcal{Z}= \mathbb{R}\times \mathbb{R}^d \) & measurement space  \\

         \( t_f, t_0 \) & Time of task termination and departure \\

         \( U(x) \) & Neighborhood of position $x$ \\
         \( p(x) \) & Probability of the gas source being located at $x$ \\
         
         \( \Phi: \mathcal{X} \to \mathbb{R} \) & Potential function  \\
         \( \tau, \boldsymbol{\tau} \) & Temperature parameter in Langevin diffusion; Temperatures of all $M$ robots $= \begin{bmatrix} \tau^{(1)}, \tau^{(2)}, \ldots, \tau^{(M)} \end{bmatrix}^T$ \\
         \( a, s(i, j) \) & Swapping rate (i.e., role-exchange probability) between robot $i$ and robot $j$ \\
         
         \( \mu, \sigma \) & Deterministic and stochastic terms in the generic stochastic differential equation \\
         \( \mathrm{d}B, \xi \) & Brownian motion in continuous and discrete time space \\
         \( \eta, h \) & Temporal and spatial discretization granularity \\
         \( \pi(\cdot) \) & Invariant distribution of MCMC sampling \\
        \bottomrule
    \end{tabular}
\end{table}

%% file: algorithm.tex
\section{Patchiness-aware Active Sensing} \label{sec:algo1}
To tackle the patchy nature of gas plumes, we design a patchiness-aware active sensing method for individual robots to enable resilient gas source localization. The strategy is inspired by Markov chain Monte Carlo (MCMC) sampling based nonconvex learning. 
In this section, we first recast the GSL problem to MCMC sampling (\cref{subsec:recast2sampling}), then we describe the algorithm design that is tailored for olfactory mobile robots sensing actively in realistic plumes (\cref{subsec:movement_strategy}). At last, we introduce the probabilistic gas source estimation method, which constitutes sampling's target distribution (\cref{subsec:estimation}). \cref{fig:SniffySquad} illustrates the workflow of the latter two sections.

\subsection{Recast GSL to MCMC Sampling} \label{subsec:recast2sampling}
In this part, we first introduce a surrogate optimization via which the problem formulated in \cref{eq:formulation} can be solved. Regarding the patchy nature of gas plumes, the surrogate optimization is then reframed as MCMC sampling. At last, we describe the Langevin MCMC, based on which we design our algorithm tailored for mobile olfactory robots. 

\subsubsection{Surrogate optimization}

Our task, gas source localization, involves identifying a position with the highest probability of locating the source of a gas emission. 
From this perspective, the objective of \cref{eq:formulation} can be realized by optimizing the likelihood of being the source location, with respect to position $x$. An efficient optimizing procedure leads to an efficient source seeking algorithm. 

Specifically, instead of minimizing the terminating time $t_f$ directly, we can optimize a surrogate function $\Phi(x):\mathcal{X} \to \mathbb{R}$, which assesses the likelihood of position $x$ being the source location (position $x$ with a small value of $\Phi(x)$ has a high probability of being the source of gas leakage). Thus, the core task can be reinterpreted, leading to the formulation: 
\begin{equation} \label{eq:min_Phi}
    \min_{x\in \mathcal{X}} \Phi(x). 
\end{equation} 
We describe the probabilistic estimator and the design of potential $\Phi(x)$ in \cref{subsec:estimation}. 

\subsubsection{Optimization via sampling}
In the context of GSL, however, the optimization task presents notable challenges attributable to two factors.
First, the potential function exhibits multimodal characteristics and is far from convex due to the patchy gas plume. Second, the potential function available is noisy and stochastic \cite{kingma2014adam}. Specifically, the robots rely on measurements obtained from onboard sensors, which are susceptible to noise and provide partial observations. 
Consequently, conventional algorithms suitable for convex objectives and methods taking deterministic movements can be trapped and/or misled. 

% Advantages of MCMC sampling
To deal with the challenges posed by the nonconvex and stochastic nature of the potential function, reframing the optimization problem to an Markov chain Monte Carlo (MCMC) sampling problem \cite{cheng2020interplay} is one promising approach. 
Monte Carlo sampling has been shown to be a powerful method for nonconvex and multimodal optimization \cite{ma2019samplingcanbefaster, chen2020accelerating, deng2020contourSGLD}, as well as noises and randomness in the potential function \cite{kingma2014adam}. 
Specifically, sampling can be faster than optimization for some nonconvex objectives \cite{ma2019samplingcanbefaster}. 
Moreover, for noisy objectives whose randomness might arise from inherent function noise, MCMC sampling algorithms can be efficient by injecting the right amount of noise into each iterative update step, providing a means of coping with inherent noise. For instance, stochastic gradient descent methods \cite{kingma2014adam} are designed to cope with such randomness. 

% Equilibrium distribution of MCMC sampliing
A key step in MCMC sampling is constructing a Markov chain that has the desired probability distribution $\pi (x)$ as its equilibrium distribution. Under this condition, one can draw samples from the intended distribution by recording states from the chain. Specifically, convergence to equilibrium means that the Markov chain 'forgets' about its initial state and the sequence of states after convergence to distribution $\pi (x)$. 

% % Relationship between sampling and optimization
% Before clarifying the rationale behind embracing a sampling-based method, we illuminate the connections between optimization and sampling \cite{cheng2020interplay}. 
% In one direction, sampling algorithms can be viewed from an optimization perspective. By sampling a probability density function 
% \begin{equation} \label{eq:invariant_distribution}
%     \pi (x)=e^{- k \Phi(x)},
% \end{equation}
% random samples drawn from $\pi (x)$ will be attracted to the peak in the distribution $\pi (x)$, i.e., the global minimum of $\Phi(x)$. This relationship is exemplified when $\pi (x) = \delta(x^{*})$, where $x^{*} = \arg\min \Phi(x)$, as $k \rightarrow +\infty$. In particular, the Dirac delta equilibrium distribution concentrates at $x^{*}$ (the global minimum of $\Phi(x)$ and hence the position with the maximum likelihood of locating the source) and the chain 'forgets' about initial position of the mobile robot. 
% % why is the SDE called sampling ?
% Conversely, optimization algorithms can be expressed by a Markov chain for sampling. Stochastic gradient descent (SGD), for instance, $\Phi(x) = \frac{1}{\eta} \sum_{i=1}^{\eta} \Phi_{i} (x)$. It aligns with the form of a sampling procedure: $x \leftarrow x- \delta \nabla \Phi(x) + \delta \xi(x,\eta)$ where the noise term $\xi(x,\eta) = \nabla \Phi(x) - \frac{1}{\eta} \sum_{i=1}^{\eta} \Phi_{i} (x)$. 

\begin{figure}[t]
    \setlength{\abovecaptionskip}{0.2cm}
    \setlength{\belowcaptionskip}{-0.3cm}
    \centering
    \includegraphics[width=1.0\columnwidth]{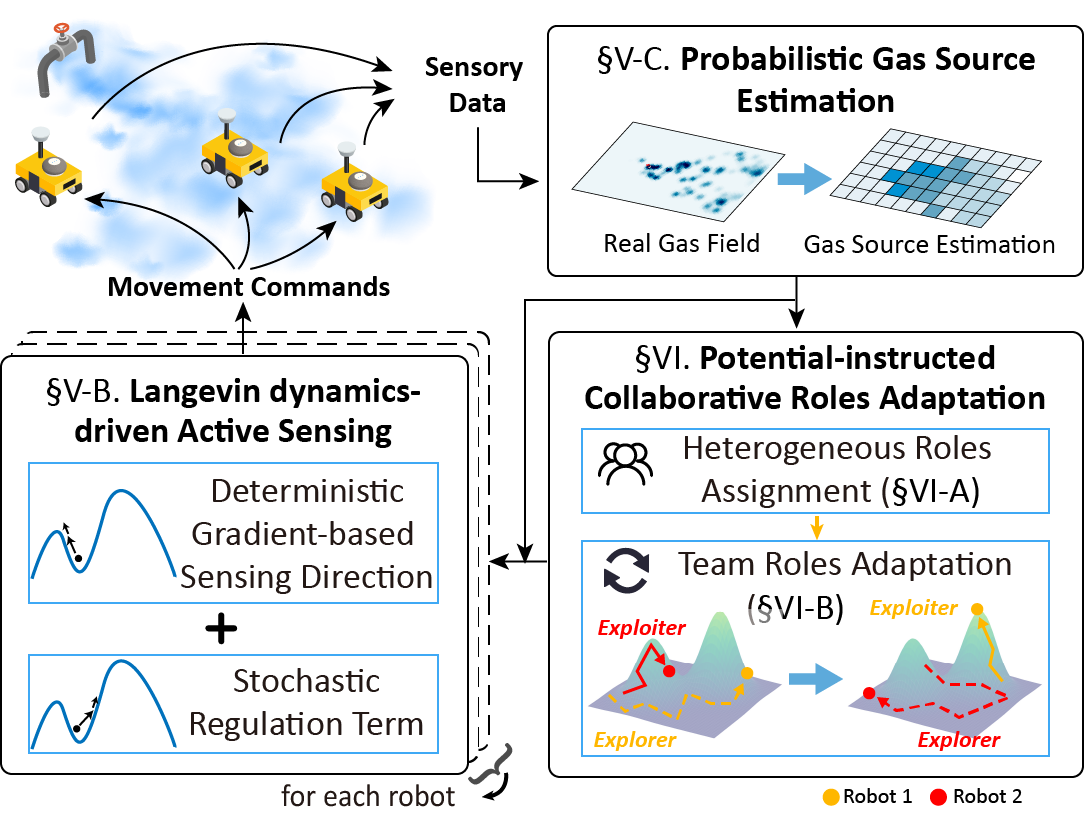}
    \caption{Illustration of the algorithm. }
    \label{fig:SniffySquad}
\end{figure}

\subsubsection{Langevin diffusion}\label{subsec:ld}
To translate MCMC sampling methods to robot movement algorithms, we design the strategy based on Langevin diffusion MCMC. 
Now, we first introduce the generic diffusion MCMC, then introduce the Langevin diffusion. 

% Generic diffusion process, SDE (discrete approximation, 
Diffusioin MCMC sampling methods simulates a Markov chain using diffusion processes, a concept from statistical physics describing the movement of particles. 
A diffusion process is defined by a stochastic differential equation (SDE), whose general form can be expressed as: 
\begin{equation}
    dx(t)= \mu(x(t)) dt+ \sigma(x(t)) dB(t), 
\end{equation}
where $x(t)$ is the positions of the particle. This SDE consists of a deterministic drift term $\mu(x(t))$ and a stochastic term, which is proportional to the standard Brownian motion $dB(t)$ and scaled by a factor $\sigma(x(t))$. 
The standard Brownian motion $\{B(t)\}_{t \geq 0}$ is a stochastic process with properties (\textit{i}) movement during any time interval $(s,t) (s<t)$, $B(t-s) \doteq B(t)-B(s)$ is a normal random variable, i.e., $B(t-s) \sim \mathcal{N}(0,t-s)$, and (\textit{ii}) the increments of different time intervals are independent.

% Langevin diffusion's SDE, intuitive description
The Langevin dynamic, with respect to the potential function $\Phi(x)$, is a diffusion process exhibiting equilibrium behaviour. It is given by a stochastic differential equation: 
\begin{equation} \label{eq:langevin_diffusion}
    dx(t)=-\nabla \Phi(x(t)) dt+\sqrt{2 \tau} dB(t),
\end{equation} 
where $\nabla \Phi(x(t))$ is the gradient of the potential function at position $x(t)$, $\{B(t)\}_{t \geq 0}$ is a standard Brownian motion and $\tau > 0$ is a parameter called temperature in statistical physics and thermodynamics. 
Notably, \cref{eq:langevin_diffusion} is the stochastic version of gradient descent, differing in the incorporation of noise term, which is Gaussian with zero mean and a covariance determined by the temporal duration of the movement. 

% Langevin diffusion‘s equilibrium: Boltzmann distribution
This random process $\{x(t)\}_{t \geq 0}$ defined in \cref{eq:langevin_diffusion} has a unique invariant distribution (a.k.a. the Boltzmann-Gibbs distribution) with density 
\begin{equation} \label{eq:boltzmann_distribution}
    \pi_\tau (x)=\frac{e^{-\Phi(x)/\tau}}{\int_{\mathcal{X}} e^{-\Phi(x)/\tau} dx} = \frac{1}{Q} e^{-\Phi(x)/\tau}.
\end{equation}
Here the denominator $Q$ is the normalization constant. With any initial distribution of $x(t)$, the limiting distribution of $X_t$ converges to $\pi_\tau (x)$  by running the diffusion process. In other words, our solution is always expected to fall into the neighborhood of a global minimum of $\Phi(x)$ with high probability.

\subsection{Langevin Dynamics-driven Active Sensing Strategy} \label{subsec:movement_strategy}

In this part, we describe the novel active sensing strategy for GSL with resiliency to patchy gas plume caused by atmospheric turbulence and noisy partial observations from robots. 

As illuminated in the previous section, gas source search and localization can be recast to drawing samples from $\pi (x)=e^{- k \Phi(x)}$, so that random samples drawn from $\pi (x)$ are attracted to the peak in the distribution $\pi (x)$, i.e., the global minimum of $\Phi(x)$. 
This relationship is exemplified when $\pi (x) = \delta(x^{*})$, where $x^{*} = \arg\min \Phi(x)$, as $k \rightarrow +\infty$. In particular, the Dirac delta equilibrium distribution concentrates at $x^{*}$ (the global minimum of $\Phi(x)$ and hence the position with the maximum likelihood of locating the source) and the chain 'forgets' about initial position of the mobile robot. 

% Gap between MCMC sampling and the robotic movement strategy
We aim to develop an active sensing strategy for source-seeking robots driven by Langevin dynamics. However, there is a gap between an algorithm tailored for robots and MCMC sampling due to the difference between particle dynamics and the mobile robots' motion characteristics. Langevin diffusion is developed to describe the motion of particles (or the update of parameters of machine learning models) by defining a continuous-time stochastic process, thus Langevin MCMC can not be directly applied. To establish strategy suitable for robots determining their sensing direction, we first discretize the diffusion process and subsequently generate a sequence of positions for mobile robots to follow. 

In the first step, we attain a discrete-time algorithm through the Finite Difference Method \cite{strikwerda2004finite}. In detail, we first discretize the states over time from $\{x(t)\}_{t \in \mathbb{R}^{+}}$ to $\{x_k\}_{k \in \mathbb{Z}^{+}}$, where $x_k=x(k \cdot \delta t)$, and $\delta t$ represents the discretization time step size. Then we apply the explicit, forward-time Euler scheme for temporal discretization, leading to the following formulation of the discrete stochastic differential equation: 
\begin{equation} \label{eq:langevin_diffusion_discrete}
    x_{k+1} = x_{k} - \eta \cdot \nabla \Phi(x_k) + \sqrt{2 \eta \tau} \cdot \xi_{k}. 
\end{equation}
Here the stepsize $\eta > 0$ is proportional to the temporal discretization granularity $\delta t$, $\tau$ is the temperature parameter, and $\{\xi_k\}_{k\in \mathbb{Z}^+}$ signifies a sequence of standard normal random vectors. \cref{eq:langevin_diffusion_discrete} describes how the robot's sensing direction evolves overtime while incorporating randomness. 

In the above equation, it requires calculating the gradient of the potential function at position $x$ to generate a sequence of trajectory points. However, the function we designed in \cref{subsec:estimation} is discrete and not differentiable. To address this, we adopt the central-difference approximation for spatial discretization: 
\[\nabla \Phi(x)\approx \begin{bmatrix} \frac{\Phi(x+h,y) - \Phi(x-h,y)}{2h}, & \frac{\Phi(x,y+h) - \Phi(x,y-h)}{2h} \end{bmatrix}^T.\] 
% $\nabla \Phi(x)\approx {[ (\Phi(x+h,y) - \Phi(x-h,y)) / 2h, (\Phi(x,y+h) - \Phi(x,y-h)) /2h ]}^T$.
Additionally, following gradients might force the robot jump between distant positions when the slope of the potential function at current position is substantial. For robots' path scheduling, we constrain the drift term to be below the robot's maximal velocity.

\subsection{Probabilistic Gas Source Estimation} \label{subsec:estimation}
With the reframing of the GSL task as an MCMC sampling problem, a definition of the potential function $\Phi(x)$, as stated in \cref{eq:min_Phi} and in equilibrium distribution \cref{eq:boltzmann_distribution}, is required. 
For every position $x\in \mathcal{X}$ in the environment, we estimate the probability that $x$ is located within the gas plume region. This probability estimation is denoted as $p(x)$, and subsequently, we define the potential function as: $\Phi(x) \doteq -log(p(x))$. 

To obtain $p(x)$ we employ a Bayesian estimation methodology, placing our method within the class of probabilistic GSL algorithms. Specifically, this estimator continually updates the likelihood of each position being situated within the gas plume region based on incoming measurements $z$ gathered from sensors equipped on the mobile robots. Mathematically, $p(x | z_{1:t}) \propto p(x | z_{1:t-1}) p(z_t | x,z_{1:t-1})$. 
In our scenario, the measurements $z=(z^c,z^w)\in \mathcal{Z}= \mathbb{R}\times \mathbb{R}^d$  collected from the mobile robots consists of the gas concentration $z^c$ and the wind vector $z^w$, acquired by chemical sensor and wind sensors (anemometers) respectively. 
To accommodate environments of arbitrary geometries, the area is discretized into a grid of equally-sized cells. 

The update procedure upon receiving new measurements is implemented based on \cite{ojeda2021information}, involving two sequential steps. 
(1) \textit{Local estimation generation}. 
% Let $U(x(t))$ denote the neighborhood region around the robot's current position $x(t)$. For positions $x \in U(x(t))$. The posterior probability of "the gas source being located at $x$" can be modeled using the wrapped normal distribution: \(p(x|z_{1:t}) = \frac{1}{\sigma \sqrt{2\pi}} \sum_{k=-\infty}^{\infty} \exp({-(\alpha(x,x(t))-\theta(z_t^w, z_t^c, x(t))+2\pi k)^2} / {2 \sigma^2})\). Here $\alpha:\mathcal{X}\times\mathcal{X} \to \Pi$ is the angle of the vector starting from $x(t)$ and pointing to position $x$. As for $\theta:\mathcal{Z} \times \mathcal{X} \to \Pi$, if $z_t^c \geq C$, it is the upwind direction; if $z_t^c < C$, $\theta(z_t^w, z_t^c) = \alpha(x(t), x(s))$ with $s$ satisfying $z_s^c \geq C$ and $\forall \tau \in (s,t], z_\tau^c <C$. 
The first step follows two intuitive and empirical rules: a) if the robot detects gas, the likelihood of neighboring grids within the upwind direction to localize within the gas plume is increased; b) If gas concentration is negligible or falls below a predefined threshold, grids directing to the position with previously recorded nonzero gas readings are attributed a higher likelihood of hosting gas. 
(2) \textit{Global estimation propagation}. 
The first step generates estimations solely for $x \in U(x(t))$, where $U(x(t))$ denotes the neighborhood region around the robot's current position $x(t)$, 
For the remaining areas $\{x \in \mathcal{X} \setminus U(x(t))\}$, their probabilities are updated in the second step by propagating information from $U(x(t))$ to the entire environment $\mathcal{X}$ based on \cite{ojeda2021information}, conforming to the geometry of the area.

\begin{algorithm}
\caption{\textit{SniffySquad}.}\label{alg:SniffySquad}
    \KwIn{Robots' initial positions $\mathrm{x}^{(0)}$}
    \KwOut{$d_\varepsilon$-approximate gas emission source position}
    \SetKwFunction{CoRA}{CollaborativeRolesAdaptation}
    \SetKwProg{Function}{Function}{}{}
    \While{$\forall {i \in \{1, 2, \ldots, M\}}, \  {\Vert x^{(i)}(t_f) - x_s \Vert} \geq d_\varepsilon$}{
        Collect environmental sensory measurements $z=(z^c,z^w)$\;
        Update belief $b(x)$, and potential $\Phi(x)$ accordingly\;
        \CoRA{}\;
        \For{ $i$ in $\{1, 2, \ldots, M\}$ }{
        Assign a role $\tau^{(i)}$ to robot $i$\;
        % Calculate $\nabla \Phi(x_k)$ using \;
        Move robot $i$ following $x_{k+1}^{(i)} = x_{k}^{(i)} - \eta \cdot \nabla \Phi(x_k^{(i)}) + \sqrt{2 \eta \tau^{(i)}} \cdot \xi_{k}$\;
        \lIf{$d(x_{k+1}^{(i)}, x_s) < d_\varepsilon$}{return}
        }
    }
    \Function{\CoRA{}}{
    \For{ $i, j$ in $\{1, 2, \ldots, M\}$ }{    
        Calculate swapping rate $s(i, j) = a \cdot \exp( \min(0, (\frac{1}{\tau^{(i)}}-\frac{1}{\tau^{(j)}}) \cdot (\Phi(x^{(i)})-\Phi(x^{(j)})))$\;
        \If{{\fontfamily{qcr}\selectfont random()} $\leq s(i,j)$}{
            Robot $i$ and $j$ swap their roles (i.e., temperatures $\tau^{(i)}$ and $\tau^{(j)}$)\;
        }
    }
    }
\end{algorithm}

\section{Potential-instructed Collaborative Roles Adaptation} \label{sec:algo2}

In this part, we present how to coordinate a team of collaborative robots to enhance source-seeking efficiency, utilizing both external environmental cues and internal team dynamics. This strategy consists of two tightly-coupled modules: \textit{Heterogeneous Roles Assignment} and \textit{Team Roles Adaptation}. 
We present our algorithm in \cref{alg:SniffySquad}. 
% %role-based collaboration
% For the purpose of locating the gas source, we propose a multi-robot motion planning algorithm, consisting of division and adaptive exchange of robot roles, thereby empowering the system through tight collaboration and systematic coordination. 

\subsection{Heterogeneous Roles Assignment} 

% Necessity of roles heterogeneity
The scattered, patchy nature of gas plumes and the multimodal nature of concentration distribution map require an GSL system with two capabilities, (\textit{i}) discovering new potential source positions and (\textit{ii}) distinguishing existing candidate positions. This gives rise to a trade-off between “global exploration” and “local exploitation" of the environment. To attain the advantages of both behaviors, we assign robots in the team with heterogeneous roles, named \textit{explorer} and \textit{exploiter} respectively. Robots acting as the former role are mainly responsible for objective (\textit{i}) while the latter focus on objective (\textit{ii}). 

% Implementation of role assignment
Considering this, we assign different roles to robots in a team. 
The role assignment can be implemented by regulating the temperature parameter $\tau$ in \cref{eq:langevin_diffusion}. Specifically, temperature $\tau$ in the motion dynamic plays a crucial part in regulating the behaviour of the robot. Robots assigned a low temperature carry out exploitative actions while those with high temperatures engage in exploratory actions. 
Mathematically, for a team with $M$ robots, consider $M$ Markov chains driven by Langevin diffusions: 
\begin{equation} \label{eq:multi_langevin_diffusion}
    \mathrm{d}x^{(i)}(t)=-\nabla \Phi(x^{(i)}(t)) \mathrm{d}t+\sqrt{2 \tau^{(i)}(t)} \mathrm{d}B^{(i)}(t),
\end{equation}
where $i \in \{1, 2, \ldots, M\}$. The role of robot $i$ is specified by its temperature $\tau^{(i)}(t)$, which governs the extent of exploring potential gas source positions and exploiting local information to distinguish spurious gas patches. When a robot is assigned a low temperature, its motion is governed by the gradient of the potential function, the robot tends to exploit the local information of the surrounding potential field. For a robot with a higher temperature, more randomness is injected into the motion, causing it to explore the environment more actively. 

% Invariant distribution
The role division module, following \cref{eq:multi_langevin_diffusion}, have a guarantee that robots are expected to concentrate around the global minima of the potential function $\Phi(x)$. Specifically, we denote the positions of $M$ robots as $\mathbf{x} = \begin{bmatrix} x^{(1)}, x^{(2)}, \ldots, x^{(M)} \end{bmatrix}^T$, and their corresponding temperature parameters as $\boldsymbol{\tau} = \begin{bmatrix} \tau^{(1)}, \tau^{(2)}, \ldots, \tau^{(M)} \end{bmatrix}^T$. The invariant joint distribution \cite{chen2020accelerating} takes the form: 
\begin{equation} \label{eq:multi_boltzmann_distribution}
    \pi_{\mathbf{\tau}}(\mathbf{x}) \propto \exp\left( - \sum_{i=1}^{M} \frac{\Phi(x^{(i)})}{\tau^{(i)}} \right).
\end{equation}
The marginal stationary distribution of the low temperature robot will be attracted to the peak of the potential function $\Phi(x)$. 

% \begin{figure}[h]
%     \setlength{\abovecaptionskip}{0.3cm}
%     \setlength{\belowcaptionskip}{-0.0cm}
%     \centering
%     \includegraphics[width=1.0\columnwidth]{figs/algo2.png}
%     \caption{The team roles adaptation algorithm.}
%     \label{fig:algo2}
% \end{figure}

\subsection{Team Roles Adaptation} 
This submodule adaptively adjusts team roles through role swapping, as illustrated in \cref{fig:SniffySquad}, based on the principles of replica exchange. 

% Necessity of roles adaptation
Optimizing the trade-off between localization effectiveness and search efficiency requires robots in the team to adjust their roles adaptively. 
One way to view this is through the lens of MCMC sampling. In \cref{eq:langevin_diffusion}, a high temperature fosters broader exploration across the global domain at the expense of reduced concentration around the minima, and vice versa. Therefore, the potential of using a fixed temperature is inherently limited. 
Alternatively, consider the gas source localization scenario. At an individual level, robots should exploit parsimoniously to avoid trapping in spurious patches and explore appropriately to escape from areas improbable to locate the source. From a team perspective, there should be some robots exploiting while others exploring. 

% Main idea of roles adaptation: swapping
In order to coordinate the team behavior systematically, we propose a role-swapping mechanism for team roles adaptation, incorporating replica exchange MCMC sampling \cite{earl2005paralleltempering}. When an exploiter $i$ locates at a position that is less likely to correspond to the desired gas source compared to an explorer $j$, this pair of robots swaps their roles. Consequently, robot $i$ transitions into an explorer so that it can turn to more promising areas and save time, while robot $j$ transforms into an exploiter who slows down and focuses to avoid missing the source of interest. 

% Implementation of swapping-based roles adaptation
Formally, this is achieved by swapping the temperatures between robots exhibiting different potential. Specifically, we consider a pair of random processes, denoted as robot $i$ and $j$, from the set of $M$ Langevin diffusions. The two robots swap temperatures with probability: 
\vspace{-0.20cm}
\begin{equation} \label{eq:re_swap_rate}
    s(i, j) = a \cdot \exp( \min(0, (\frac{1}{\tau^{(i)}}-\frac{1}{\tau^{(j)}}) \cdot (\Phi(x^{(i)})-\Phi(x^{(j)}))).
\end{equation}
Here the constant $a \geq 0$ represents the swapping intensity. 
% exemplify the swapping rule
In this manner, robots occupying positions associated with smaller potential function values are inclined to be assigned a role of exploiter, specified by a lower temperature, and vice versa. 
To elucidate the exchange rule, let us assume, without loss of generality, that $0 < \tau^{(i)} < \tau^{(j)} < \infty$ and $a = 1$. 
If $\Phi(x^{(i)}(t)) > \Phi(x^{(j)}(t))$, then robot $i$ swaps its temperature with robot $j$; otherwise, they swap with a probability $s(i, j) \in (0,1)$, which diminishes monotonically with respect to the difference between two objective values. For example, as $\Phi(x^{(i)}) \rightarrow 0$ and $\Phi(x^{(j)}) \rightarrow +\infty$, the swapping rate $s \rightarrow 0$. 

% Acceleration effect
Under the specific swapping rule defined in \cref{eq:re_swap_rate}, the algorithm exhibits an acceleration effect in concentrating around the global minima of $\Phi(x)$. This effect is theoretically analysed and demonstrated in \cite{chen2020accelerating}. 
Meanwhile, sampling in this manner still leads to the minimization of potential function and hence the identification of the most probable gas source location. This is because the equilibrium distribution of the replica exchange Langevin diffusion is also \cref{eq:multi_boltzmann_distribution}, the same as algorithms in which robots follow independent diffusion processes in \cref{eq:multi_langevin_diffusion}.

%% file: evaluation.tex
\begin{figure}[t]
    \setlength{\abovecaptionskip}{0.3cm}
    \setlength{\belowcaptionskip}{-0.0cm}
    \centering
    \includegraphics[width=1.0\columnwidth]{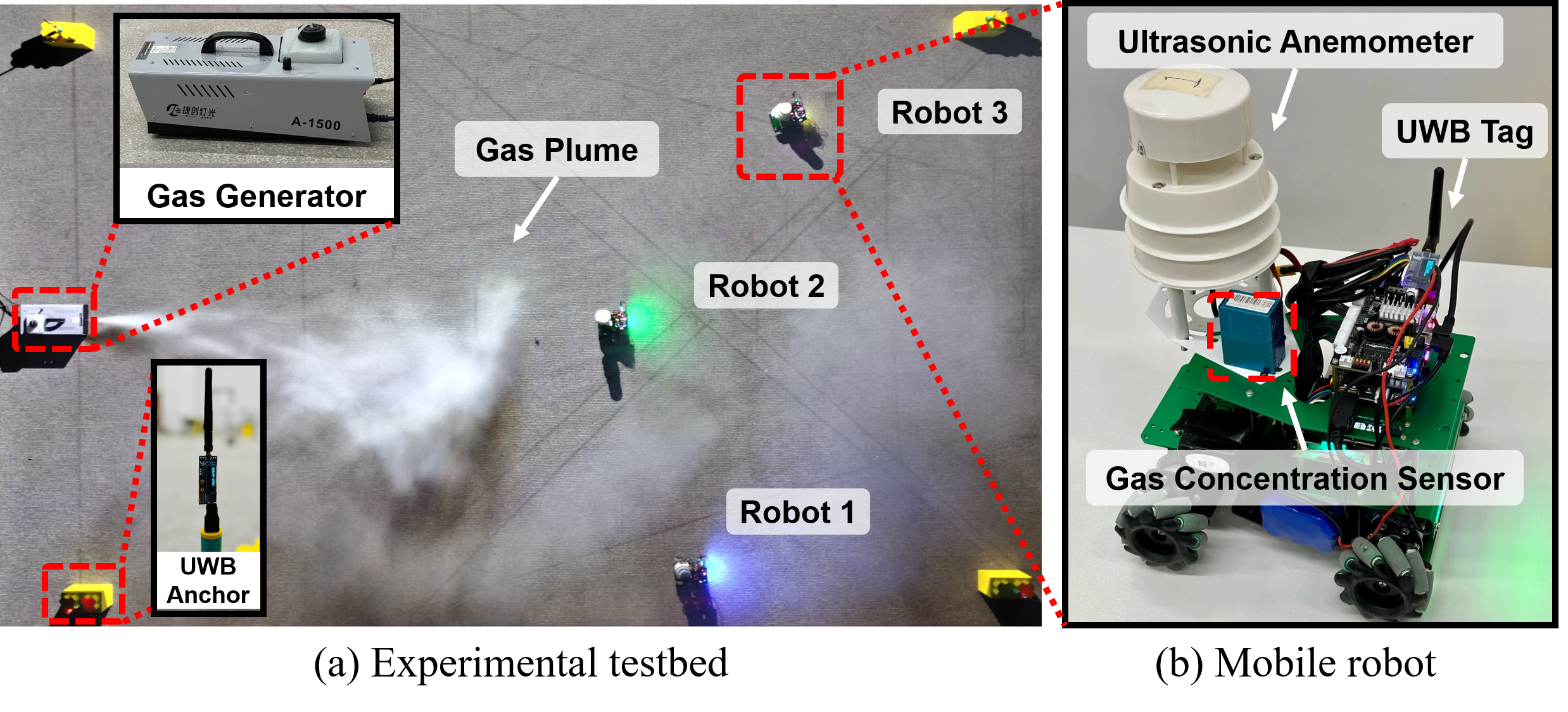}
    \caption{Experimental testbed of \textit{SniffySquad}.}
    \label{fig:testbed}
\end{figure}

\section{IMPLEMENTATION AND EVALUATION} \label{sec:eval}

% In this section we conduct extensive experiments to comprehensively evaluate the performance and advantages of \textit{SniffySquad}. 
In this section, we evaluate the system's performance from two aspects: source localization effectiveness and search efficiency.
We first describe the settings of both physical and simulation experiments, as well as the metrics used in the evaluation (\cref{subsec:experiment_setting}). We then compare overall performance of \textit{SniffySquad} in both a testbed and simulation (\cref{subsec:eval_end2end}). We also carry out experiments under different conditions to evaluate \textit{SniffySquad}'s robustness to both external and internal factors (\cref{subsec:eval_robustness}). 
% Finally, we perform an ablation study to further clarify each component (\cref{subsec:eval_ablation}). 

%%%%%%%%%%%%%%%%%%%%%%%%%%%%%%%%%%%%
\subsection{Implementation and Setting} \label{subsec:experiment_setting}

\subsubsection{Field Experiment Setup}
We first describe how we build an environment with gas, then mobile robots deployed in the testbed.

\noindent \textbf{Gas environment}. 
% As illustrated in Fig. \ref{fig:testbed}, we build a gas leakage testbed in a rectangular $15m \times 10m$ laboratory environment with turbulent wind outside. 
As illustrated in Fig. \ref{fig:testbed}, we build a gas leakage testbed in a rectangular $15m \times 10m$ laboratory environment.
% Two experimental environment geometries are elaborately set with diverse obstacle configurations. 
Since generating gas plumes for conducting field experiments requires strict safety and environmental considerations, we simulate the gas leakage scenario by adopting a smoke machine as the gas generator to release gas following previous work \cite{bourne2019coordinated, hutchinson2019experimental, hutchinson2018information}. The primary component of the smoke fluid is ethylene glycol (EINECS No. 246-770-3). 
The behavior of such plumes mirrors that of numerous gas plumes in real-world application scenarios \cite{kadakia2022odour, francis2022gas}. We deploy a fan behind the gas generator to generate turbulent wind.

\noindent \textbf{Multi-robot platform}. 
We implement \textit{SniffySquad} on commercial unmanned ground vehicles, as shown in \fig \ref{fig:testbed}(b). Each robot is equipped with a particle matter sensor Plantower PMS5003T \cite{levy2018field}, which is capable of measuring concentration of particles with diameters less than 3 micrometers, expressed in units of micrograms per cubic meter ($\mu g/m^3$). When ethylene glycol is atomized, it forms aerosol particles. The particle matter sensor is utilized to detect these particles, allowing us to measure the concentration of ethylene glycol. Additionally, an ultrasonic anemometer has been integrated to provide data on wind direction and speed. 
To establish ground truth positions, a UWB localization device is integrated. Robot movement is guided by a PID controller. 
The onboard computer of the robot is a Raspberry Pi 4 Computer Model B with 8GB RAM, and the edge server is equipped with Intel(R) Core(TM) i7-11700 of 2.50GHz main frequency and 16G RAM, running the Ubuntu 20.04.4 operating system. 
Mobile robots depart from the opposite side of the gas source, i.e., the right side of \fig \ref{fig:testbed}(a).

\begin{figure*}%[b]
\setlength{\abovecaptionskip}{0.1cm}
\setlength{\belowcaptionskip}{-0.1cm}
\centering
    \includegraphics[width=1.9\columnwidth]{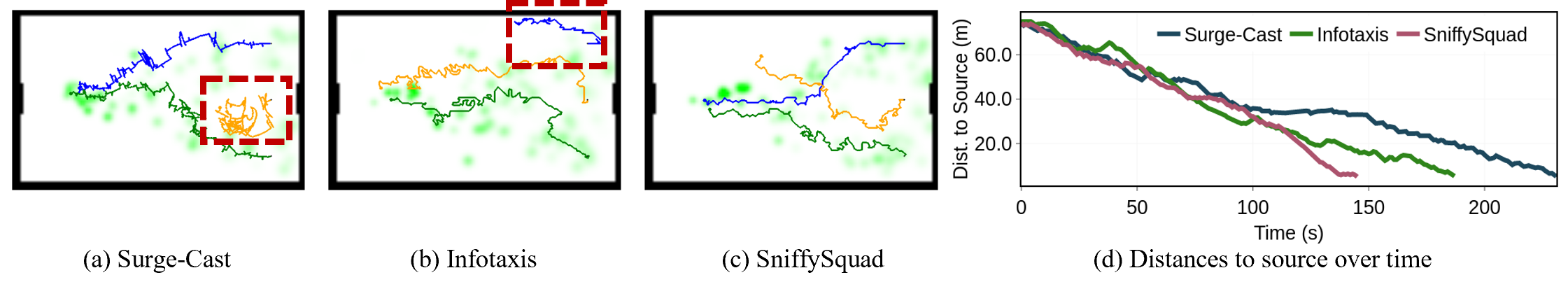}
\caption{In an empty space (w/o walls and obstacles). (a)-(c) Illustration of trajectories generated by Surge-Cast, Infotaxis, and \textit{SniffySquad}. The red rectangles indicate the trajectories that were misled by the gas patches. (d) The distance of the robot to the gas release source over time. }
\label{fig:overall_illustration-B}
\end{figure*}

\begin{table}[t]
\caption{Overall performance of field experiment.}
\label{tab:field_overall}
\centering
\setlength{\tabcolsep}{4mm}{
\begin{tabular}{ccc}
\hline
           & Success Rate & Path Efficiency(\%)    \\
\hline
SniffySquad  & \textbf{5/6}          & \textbf{90.0}            \\
Infotaxis  & 4/6          & 62.0                     \\
Surge-Cast & 4/6          & 61.0                     \\
\hline
\end{tabular}
% \begin{tabular}{cccc}
% \hline
%            & Success Rate & Path Efficiency(\%)   & Search Time(s)       \\
% \hline
% SniffySquad  & 5/6          & \textbf{90}           & \textbf{40} \\
% Infotaxis  & 4/6          & 62                    & 54          \\
% Surge-Cast & 4/6          & 61                    & 107          \\ 
% \hline
% \end{tabular}
}
\end{table}

\begin{table}[b]
\setlength{\abovecaptionskip}{0.1cm}
\setlength{\belowcaptionskip}{0.0cm}
    \caption{Configuration of gas dispersion parameters}
    \label{tab:simulation_params}
    \centering
    \small
    \begin{tabular}{ccl}
        \toprule
        \textbf{Symbol} & \textbf{Value} & \textbf{Description} \\
        \midrule
        $\mu$  & $10\ ppm$ & Gas conc. at filament center   \\
        $\sigma$ & $10\ cm$ & Initial std of filament \\
        $\gamma$ & $10\ cm^2/s$ & Growth ratio of $\sigma$  \\
        P, T              & 1 Atm, 298 Kelvins & Pressure and temperature  \\
        $\upsilon$ & $1.529\times 10^{-5} m^2/s$ & Kinematic viscosity \\
        $\rho$ & $1.196 kg/m^3$ & Air density \\
        $k$ & $3.75\times 10^{-3} m^2/s^2$ & Turb. kinetic energy \\
        $\varepsilon$ & $1.25\times 10^{-2} m^2/s^3$ & Dissipation rate \\
        \bottomrule
    \end{tabular}
\end{table}

\subsubsection{Simulation Setting}
% We build up the simulation environment based on two tools: (i) OpenFOAM \cite{jasak2009openfoam}, a computational fluid dynamics (CFD) simulator, is used to generate the wind flow vector field; (ii) GADEN \cite{monroy2017gaden}, a 3-D gas dispersion simulator widely recognized and commonly used for robotic olfaction, is used to generate the gas concentration field. Specifically, the data generation process involves the following two steps. First, OpenFOAM models the flow of wind based on the environment geometry that is configured by a computer-aided design (CAD) model, and the specified flow inlets and outlets. Second, based on the wind flow data, GADEN takes the environment CAD model as input and obtains 3D gas distributions by implementing the filament gas dispersion theory \cite{farrell2002filament}. The data generation procedure follows previous work \cite{ojeda2021information} and our parameter configuration is listed in \cref{tab:simulation_params}. 
% The geometry of environment used in simulation includes two representative scenarios in industrial factories: ($i$) empty spaces \cref{fig:geo_empty} and ($ii$) environments with rooms featuring walls and obstacles \cref{fig:geo_room}. 
To simulate gas dispersion processes, we follow previous works~\cite{ojeda2021information} and build up the simulation environment based on two tools: ($i$) OpenFOAM \cite{jasak2009openfoam}, a computational fluid dynamics (CFD) simulator, is used to generate the wind flow vector field; ($ii$) GADEN~\cite{monroy2017gaden}, a 3-D gas dispersion simulator widely recognized and commonly used for robotic olfaction, is used to generate the gas concentration field. The gas simulation process involves two key steps: First, OpenFOAM models the flow of wind based on the environment geometry that is configured by a computer-aided design (CAD) model, and the specified flow inlets and outlets; Second, based on the wind flow data, GADEN takes the environment CAD model as input and obtains 3D gas distributions by implementing the filament gas dispersion theory~\cite{farrell2002filament}. The specific parameter configuration is listed in~\cref{tab:simulation_params}. 
Our experiments include two representative scenarios in industrial factories: ($i$) empty spaces \cref{fig:geo_empty} and ($ii$) environments with rooms featuring walls and obstacles \cref{fig:geo_room}.
% \com{how to simulate the robot?}
To simulate robot's dynamics, we implement a controller based on \cite{chen2024adaptive}.

\begin{figure}
\setlength{\abovecaptionskip}{0.0cm}
\setlength{\belowcaptionskip}{-0.3cm}
\centering
    \subfigure[Success Rate]{
        \centering
        \includegraphics[width=0.48\columnwidth]{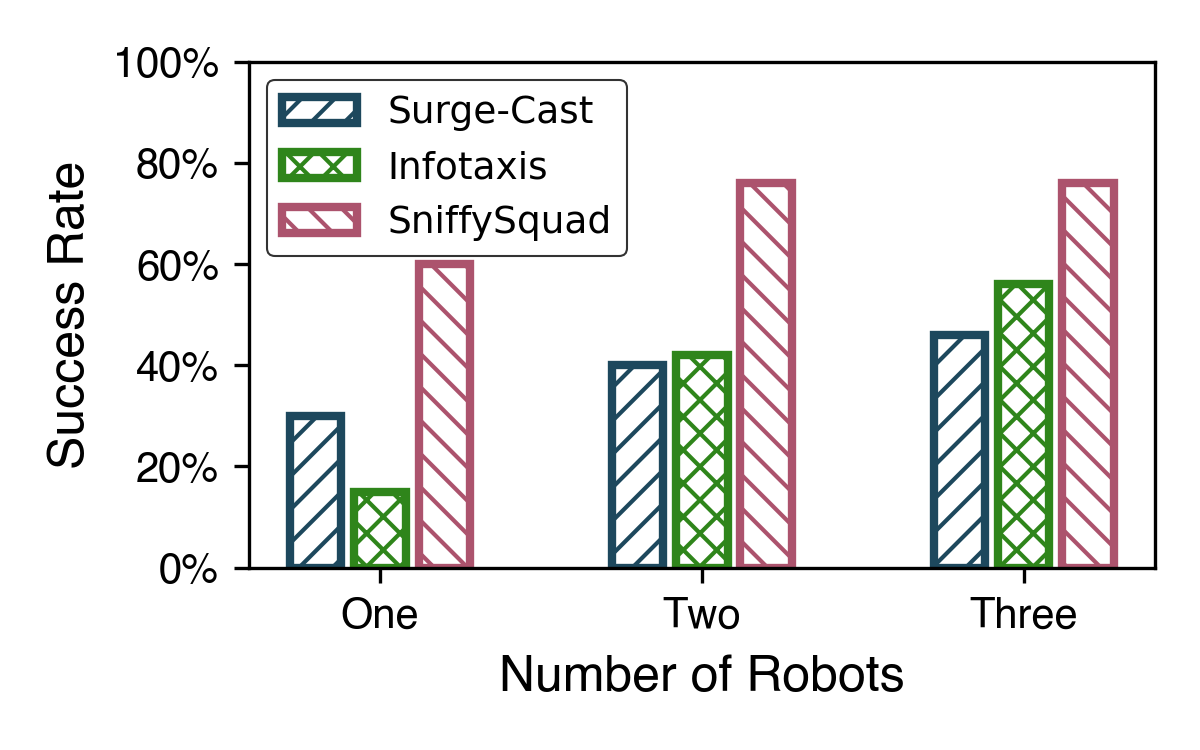}
        %\caption{fig1}
        \label{fig:numrobot_successrate}
    }%
    % \subfigure[Search Time]{
    %     \centering
    %     \includegraphics[width=0.66\columnwidth]{figs/numrobot_searchtime.png}
    %     \label{fig:numrobot_searchtime}
    % }%
    \subfigure[Path Efficiency]{
        \centering
        \includegraphics[width=0.48\columnwidth]{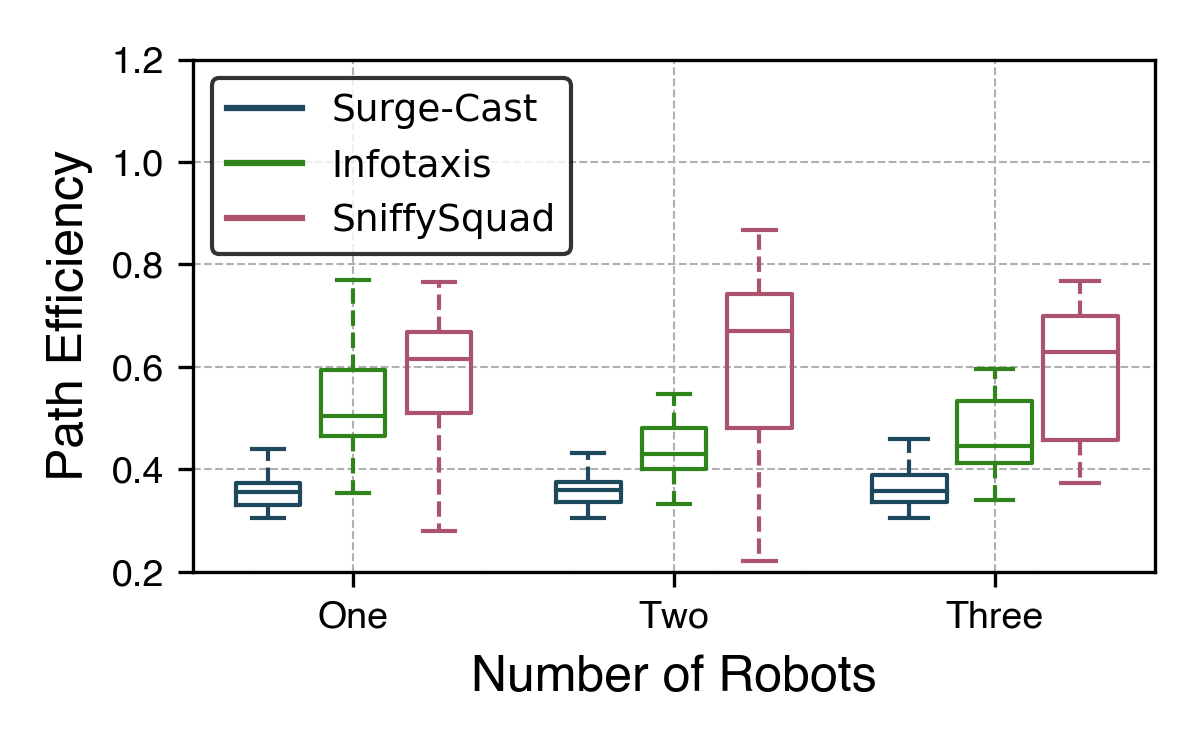}
        \label{fig:numrobot_pathefficiency}
    }%
% \caption{Overall performance of \textit{SniffySquad} and related works evaluated using different metrics.}
\caption{Overall performance of physical feature-based simulation experiments.}
\label{fig:overall_metrics}
\end{figure}

\subsubsection{Comparative Methods}
We compared \textit{SniffySquad} with two related state-of-the-art (SOTA) methods including: 
(\textit{i}) \textbf{Surge-Cast} \cite{reddy2022olfactory}, a SOTA reactive bio-inspired GSL algorithm. This algorithm involves two distinct motions patterns: when the robot identifies the plume concentration exceeding a predefined threshold, it engages in a "surge", i.e., moving upwind; otherwise, it executes a "cast" maneuver perpendicular to the wind direction to re-establish contact with the plume;
(\textit{ii}) \textbf{Infotaxis} \cite{ojeda2021information}, a SOTA probabilistic source parameter estimation-based GSL algorithm. It's an information-driven path planning algorithm, leveraging a probability map of the source position estimation and driving the robot's movement toward the place with the highest information gain.

\subsubsection{Evaluation Metrics}
We employ the following two metrics to evaluate the algorithms' effectiveness and efficiency, respectively:
\noindent (\textit{i}) \textbf{Success Rate (SR)}. It is determined by calculating the ratio of successful trials across all repeated experiments. A trial is deemed successful if mobile olfactory robots reach the vicinity of the actual ground truth position of the gas source within the predefined time limit and distance threshold $\varepsilon_s$. The threshold is set to $0.5m$ in the experiments. In practical application scenarios, the gas leakage source can be detected using onboard cameras as soon as it comes within a robot's line-of-sight. 
% \noindent (ii) \textbf{Search Time}. In the case of successful trials, the search time is defined as the time from the moment of departure to the instance of arrival at the source position. 
\noindent (\textit{ii}) \textbf{Path Efficiency (PE)}. It is defined as $\frac{d_{min}}{d}$, where $d_{min}$ is the shortest path distance from the robot's starting point to the gas source, and $d$ is the length of the robot's travel trajectory driven by GSL approaches. This metric is only calculated for successful trials. 
Note that path efficiency is proportional to the time required for localizing the gas source, which aligns with the objective in \eq \ref{eq:formulation}.
% \rev{Note that since the robots' velocities remain constant, maximizing path efficiency is equivalent to minimize the objective in \eq \ref{eq:formulation}.}

\begin{figure*}[tp]
\setlength{\abovecaptionskip}{0.2cm}
\setlength{\belowcaptionskip}{-0.3cm}
    % \begin{minipage}[b]{0.9\columnwidth}
    \centering
        \subfigure[Success Rate]{
            \centering
            \includegraphics[width=0.6\columnwidth]{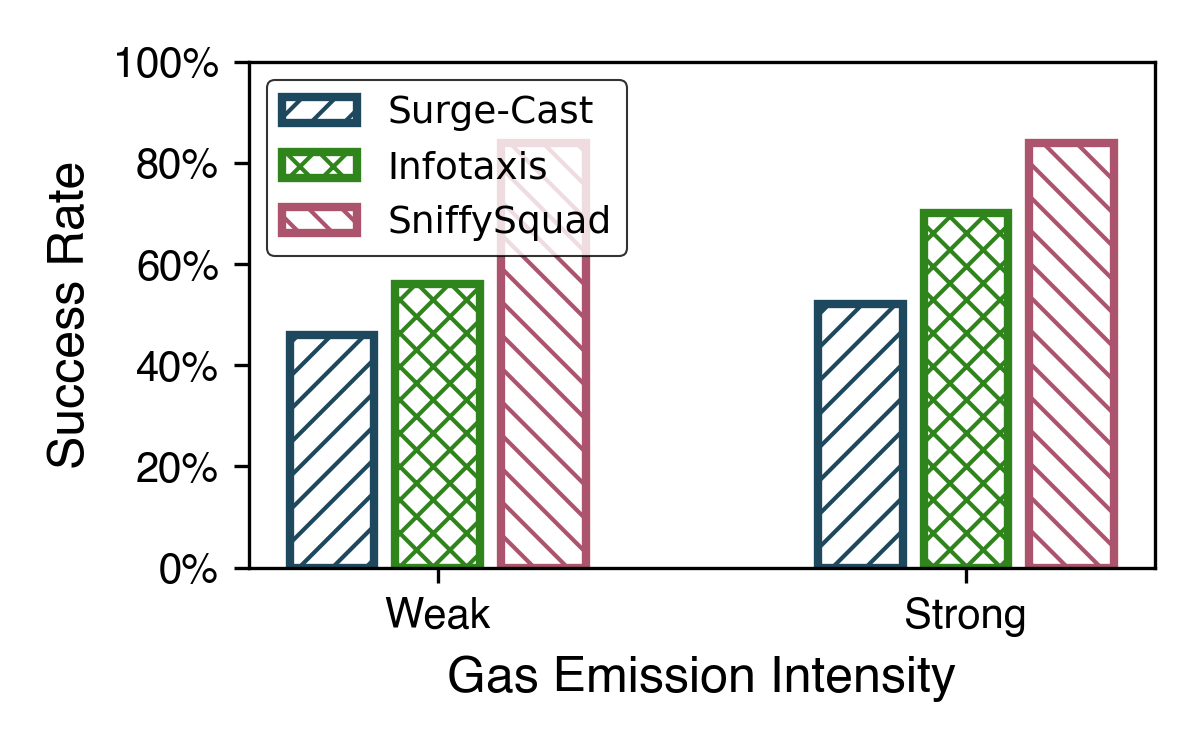}
            %\caption{fig1}
            \label{fig:Q_SR}
        }%
        \subfigure[Path Efficiency (Weak Source)]{
            \centering
            \includegraphics[width=0.6\columnwidth]{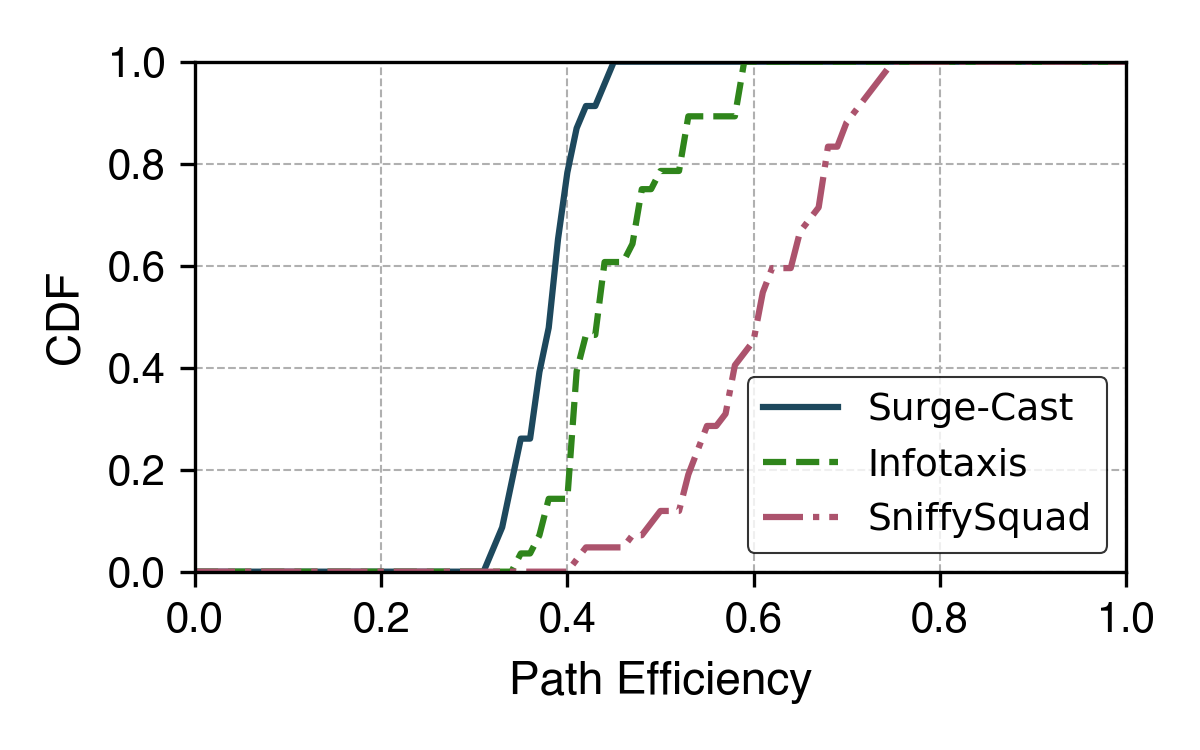}
            \label{fig:Q_E_Weak}
        }%
        \subfigure[Path Efficiency (Strong Source)]{
            \centering
            \includegraphics[width=0.6\columnwidth]{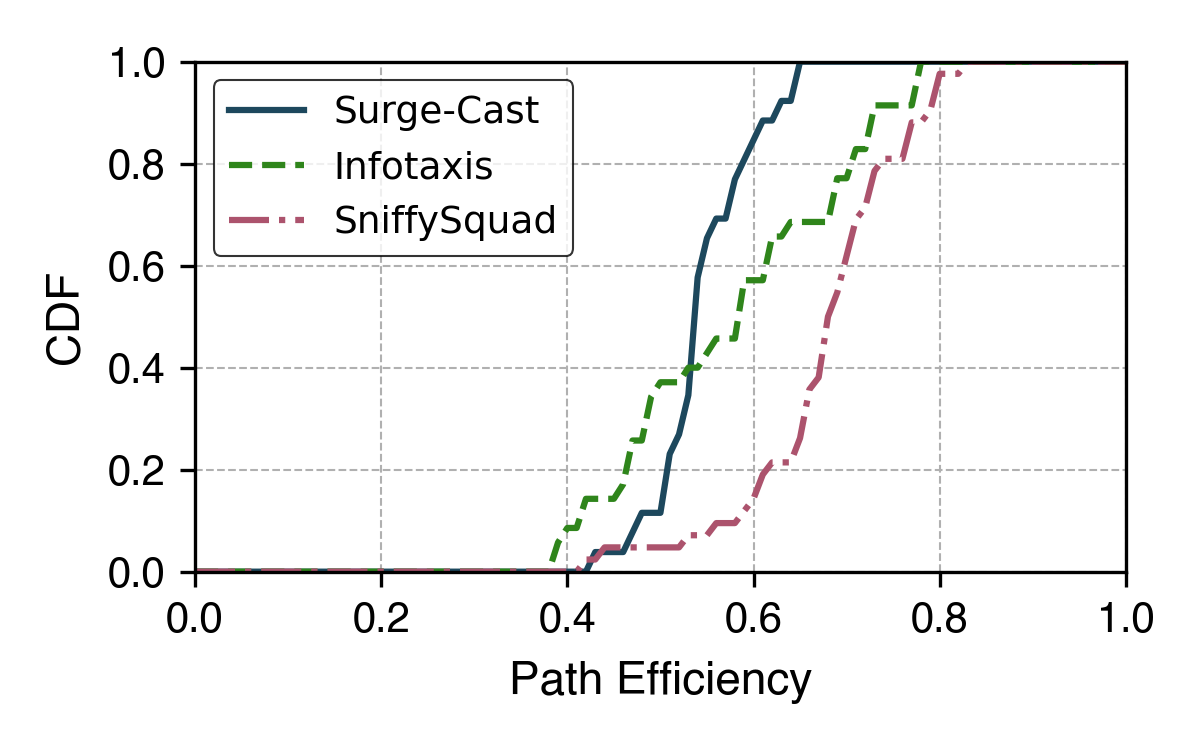}
            \label{fig:Q_E_Strong}
        }%
    \caption{Impact of gas emission intensity. (a) Success rate. (b)-(c) CDF of path efficiency under a weak and a strong source, respectively. }
    \label{fig:Q}
\end{figure*}

\begin{figure*}[tp]
\setlength{\abovecaptionskip}{0.2cm}
\setlength{\belowcaptionskip}{-0.3cm}
    % \begin{minipage}[b]{0.9\columnwidth}
    \centering
        \subfigure[Success Rate]{
            \centering
            \includegraphics[width=0.6\columnwidth]{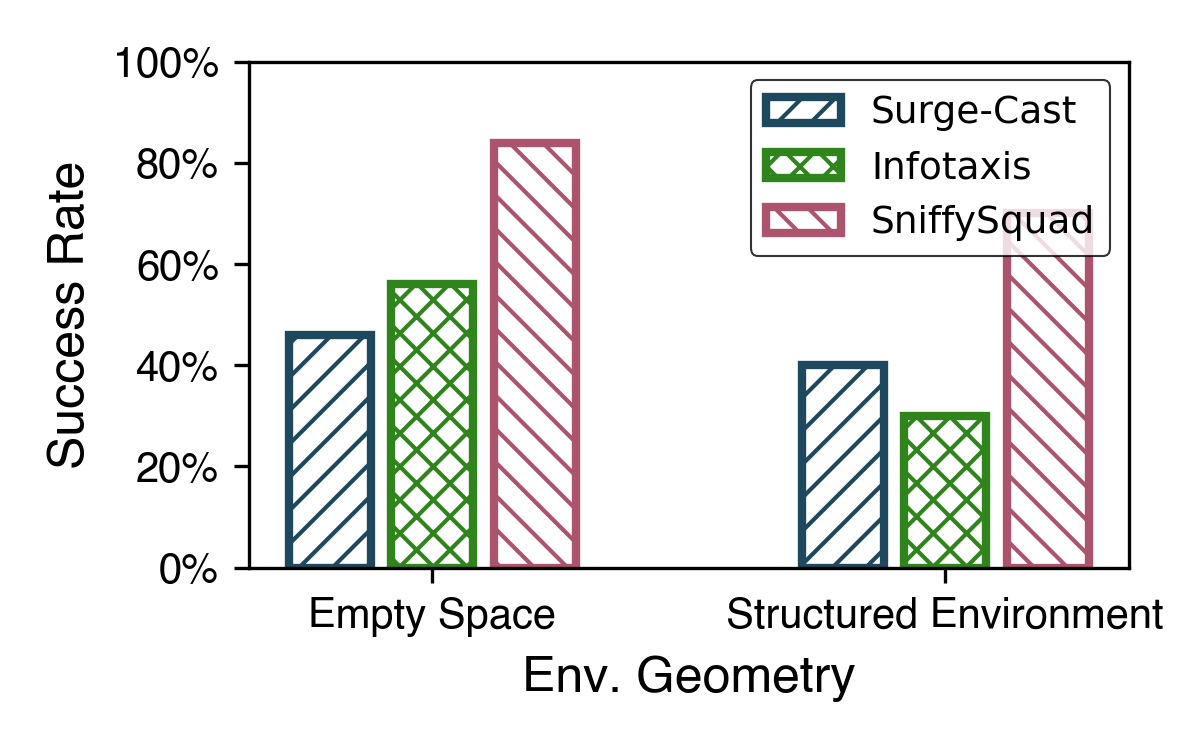}
            %\caption{fig1}
            \label{fig: geo_SR}
        }%
        \subfigure[Path Efficiency (Empty Space)]{
            \centering
            \includegraphics[width=0.6\columnwidth]{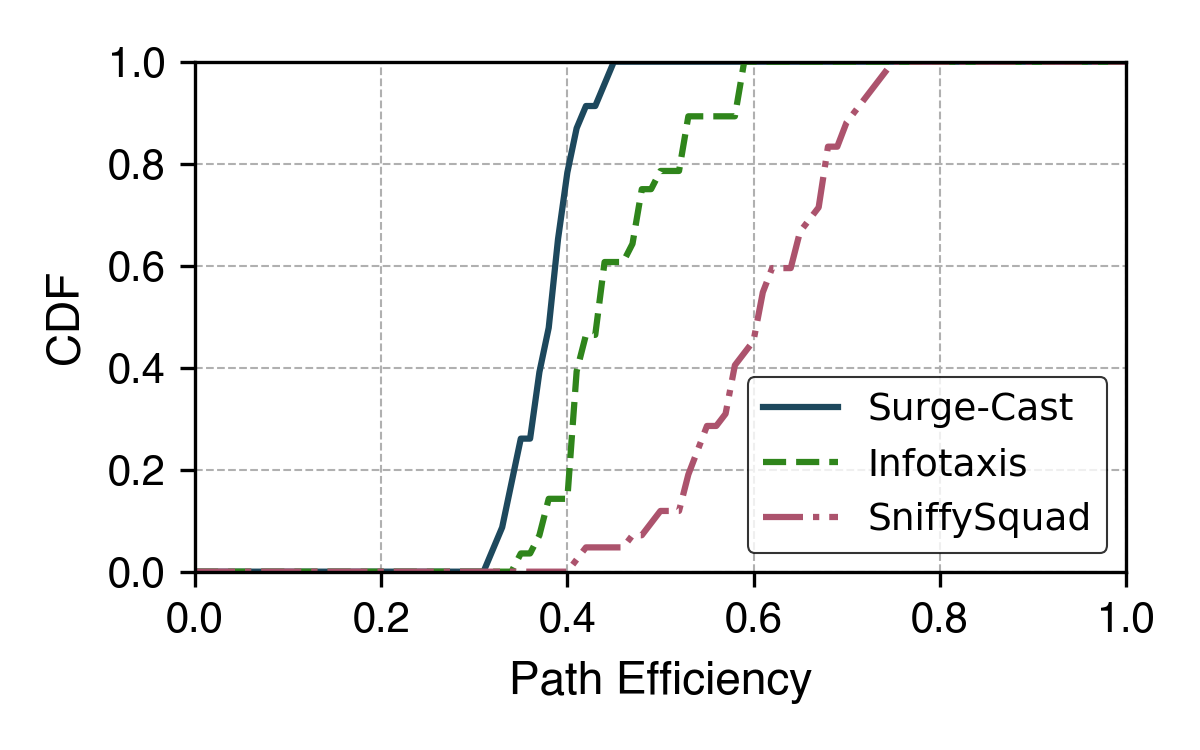}
            \label{fig: geo_E_Empty}
        }%
        \subfigure[Path Efficiency (Structured Space)]{
            \centering
            \includegraphics[width=0.6\columnwidth]{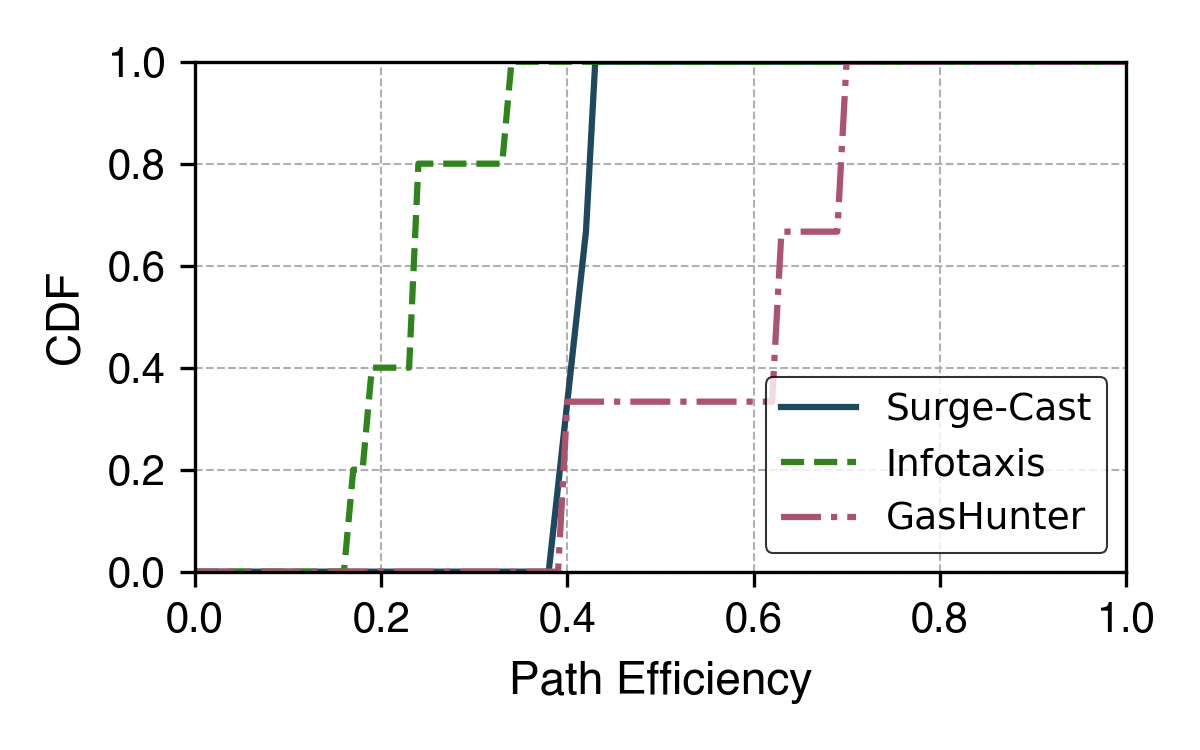}
            \label{fig: geo_E_Struc}
        }%
    \caption{Impact of environment geometry. (a) Success rate. (b)-(c) CDF of path efficiency under an empty and a structured space, respectively. }
    \label{fig: geo}
\end{figure*}

\subsection{Overall Performance} \label{subsec:eval_end2end}

% \subsubsection{Physical experiments.}
\cref{tab:field_overall} illustrates overall performance evaluated in the testbed.
A total of 6 experiments were conducted to compare the performance between Surge-Cast to Infotaxis and \textit{SniffySquad}. 
In terms of both success rate and path efficiency, the performance of \textit{SniffySquad} surpasses two SOTA baselines, Surge-Cast and Infotaxis. \textit{SniffySquad} improves the path efficiency (i.e., decreases the trajectory lengths) by $32\%$ and $32\%$ compared to Surge-Cast and Infotaxis, respectively. % \cref{fig:field_d2src_cdf} \textbf{XXX}

% In terms of both path efficiency and search time, the performance of \textit{SniffySquad} surpasses two SOTA baselines, Surge-Cast and Infotaxis. \textit{SniffySquad} improves the path efficiency (i.e., decreases the trajectory lengths) by $32\%$ and $32\%$ compared to Surge-Cast and Infotaxis, respectively; and it reduces search time by $67.5 s$ and $14 s$, respectively. 

% We also compare the results between field experiments and physical-feature-based simulation, and both sets of results exhibit similar results among the three methods. Specifically, \textit{SniffySquad} improves the path efficiency by $39\%$ and $28\%$ and reduces search time by $20.6\%$ and $15.2\%$, respectively. This similarity verifies that \textit{SniffySquad}, which aims to improve the search efficiency by explicitly considering the patchiness of the gas plumes in real world, brings significant benefits to mobile robot olfaction-based gas source localization. 

% \subsubsection{Simulation experiments.}

% The overall performance of \textit{SniffySquad} is further evaluated extensively by simulation in two representative scenarios in industrial factories: (i) an open space (measuring $100m$ by $60m$) and (ii) an area with the existence of walls, rooms and obstacles ($100m$ by $100m$). We compare the performance of \textit{SniffySquad} as well as two state-of-the-art algorithms, Surge-Cast and Infotaxis. 

\cref{fig:overall_metrics} evaluates overall performance of \textit{SniffySquad} in physical feature-based simulations, using various numbers of robots. These experiments were carried out 50 times for each configuration. 
% 1
Across all scenarios involving robot counts ranging from one to three, it is evident that \textit{SniffySquad} consistently attains the highest success rate, achieves the shortest search time for both an individual robot and robot teams, and generates the most efficient path. Specifically, \textit{SniffySquad} achieves a success rate of 76\% for a team of three robots, improving by $30\%$ and $20\%$ in comparison with Surge-Cast and Infotaxis, respectively. The search time and path efficiency are $148 s$ and $0.67$, reducing the total search time by $22.92\%$ and $18.46\%$, shortening the trajectory length by $41.79\%$ (from $2.56 d_{min}$ to $1.49 d_{min}$) and $32.84\%$ (from $2.22 d_{min}$ to $1.49 d_{min}$), respectively. This results from the resilience brought by the individual-level movement planning and the team-level roles adaptation algorithm, which enable robots to escape from false positive sources and parallelize the exploration of the environment and exploitation of collected information in a flexible manner. 
% 2
For all algorithms, success rates show an upward trend as the number of robots employed for gas source search and localization increases. Apparently, this is because the area is explored to a greater extent when more robots are working. 

To further compare the performance of \textit{SniffySquad} and baseline methods, we visualize the trajectories of all methods in \cref{fig:overall_illustration-B} (a)-(c). It is demonstrated that \textit{SniffySquad} produces smoother trajectories than baselines. Surge-Cast, a reactive method relying solely on current observations to make decisions, yields inefficient zig-zag paths and is the most susceptible to spatial and temporal gas concentration fluctuations (as shown in \cref{fig:conc_distr_characteristic}). Infotaxis, though incorporating a Bayesian gas source position estimator to filter the sensory measurements, suffers from trap in patchy areas. 
\cref{fig:overall_illustration-B} (d) illustrates the nearest distance to the source within all robots over time. It's verified that \textit{SniffySquad} navigates towards the source the most efficiently and arrives at the source the most quickly. Surge-Cast takes $231$ seconds and Infotaxis takes $187$ seconds in total while \textit{SniffySquad} takes only $145$ seconds, indicating a decrease in search time by $37.23\%$ and $22.46\%$ in comparison to the two baselines, respectively. The acceleration effect produced by \textit{SniffySquad} relative to previous methods in localizing the source attributes to the movement planning method that regulates the robots' moving direction to enable escaping from gas patches, as well as the collaborative assignment and adaptation of team roles.

\subsection{System Robustness} \label{subsec:eval_robustness}
We further experimentally evaluate \textit{SniffySquad} with respect to various environmental conditions and key parameter settings in our system.
\subsubsection{Impact of Gas Emission Intensity} 
\begin{figure}
\setlength{\abovecaptionskip}{0.2cm}
\setlength{\belowcaptionskip}{-0.3cm}
    % \begin{minipage}[b]{0.9\columnwidth}
    \centering
        \subfigure[Weak source]{
            \centering
            \includegraphics[width=0.425\columnwidth]{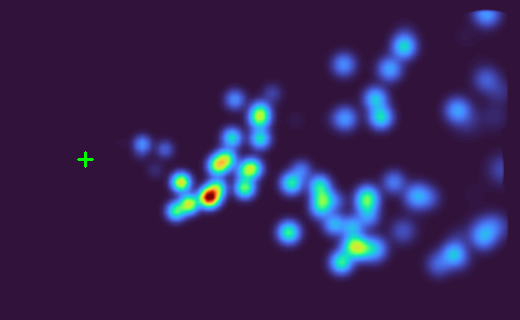}
            %\caption{fig1}
            \label{fig:Q_illustration-weak}
        }%
        \subfigure[Strong source]{
            \centering
            \includegraphics[width=0.495\columnwidth]{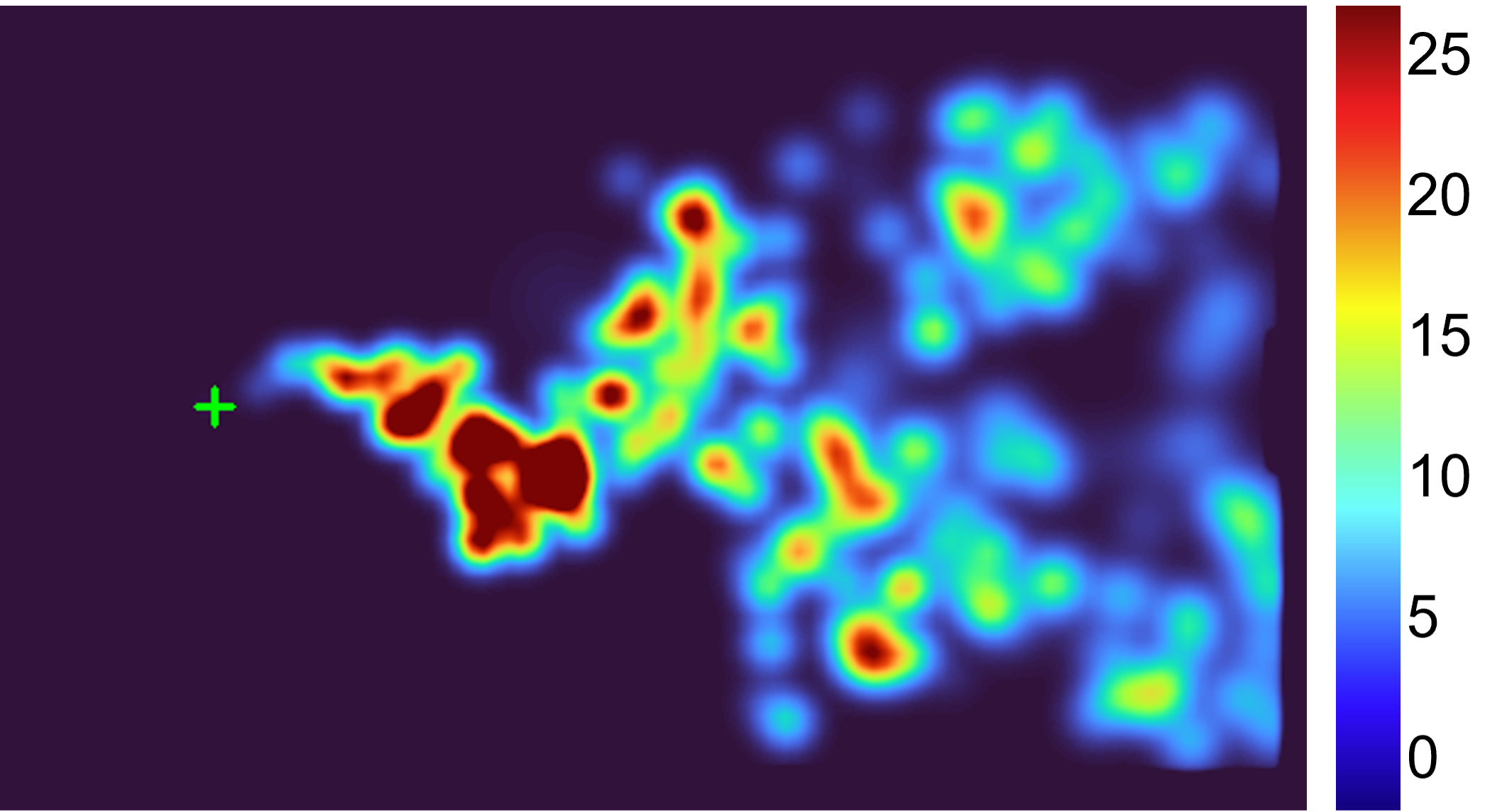}
            \label{fig:Q_illustration-strong}
        }%
        % \subfigure[Success Rate]{
        %     \centering
        %     \includegraphics[width=0.5\columnwidth]{figs/Q_successrate.png}
        %     %\caption{fig1}
        %     \label{fig:Q_SR}
        % }%
        % \subfigure[Path Efficiency]{
        %     \centering
        %     % \includegraphics[width=0.5\columnwidth]{figs/Q_pathefficiency_cdf.png}
        %     \includegraphics[width=0.5\columnwidth]{figs/Q_pathefficiency_cdf.jpg}
        %     \label{fig:Q_E}
        % }%
    \caption{Illustration of gas concentration fields under different emission intensities. }
    \label{fig:Q_illustration}
\end{figure}

In real-world scenarios, \textit{SniffySquad} should be capable of handling situations where different gas emission intensities are presented. Therefore, we investigate the performance of our system when handling different release rates of the gas source. Specifically, we select two representative gas release rates of 10 filaments per second and 30 filaments per second, which are denoted as a weak source and a strong source, respectively. As illustrated in \cref{fig:Q_illustration-weak,fig:Q_illustration-strong}, a snapshot of the field with the weak source shows a more discontinuous gas concentration field compared to that of the strong source, which indicates a higher difficulty for gas source localization. 

We compare the performance of the proposed method and baselines under these two scenarios, as shown in \cref{fig:Q}. 
% 1
Firstly, across scenarios of a strong and a weak source, our method consistently achieves the highest success rate and the most efficient paths among all methods. This can be attributed to the patchy plume-resilient movement planning mechanism, as well as role adaptation based collaboration among robots in the team. 
% 2
Secondly, for all the methods, the success rate and path efficiency degrade as the gas emission intensity from strong turns to weak. 
% Specifically, Surge-Cast, Infotaxis, and our approach's success rates decrease from $52\%$, $70\%$, and $84\%$ to $46\%$, $56\%$, and $83\%$, respectively. 
This is because the olfactory stimulus in a field induced by the weaker source is sparser, confusing the robots more severely.
% 3
Thirdly, it's noteworthy that our approach's advantage is especially significant under the weak gas source, validating its remarkable robustness to gas emission intensity. From the strong to weak source scenario, Surge-Cast and Infotaxis' success rates decrease considerably by $12\%$ and $20\%$, respectively, while their path efficiencies degrade significantly by $41\%$ and $33\%$, respectively. In contrast, \textit{SniffySquad} still maintains a success rate of above $89\%$ and its path efficiency slightly reduces by only $13\%$. This is because of the integrated design of sensing and planning processes in our method, which allows the robot to escape from spurious gas patches and leads to more efficient localization.

\begin{figure}
\setlength{\abovecaptionskip}{0.2cm}
\setlength{\belowcaptionskip}{-0.3cm}
    % \begin{minipage}[b]{0.9\columnwidth}
    \centering
        \subfigure[Empty Space]{
            \centering
            \includegraphics[width=0.49\columnwidth]{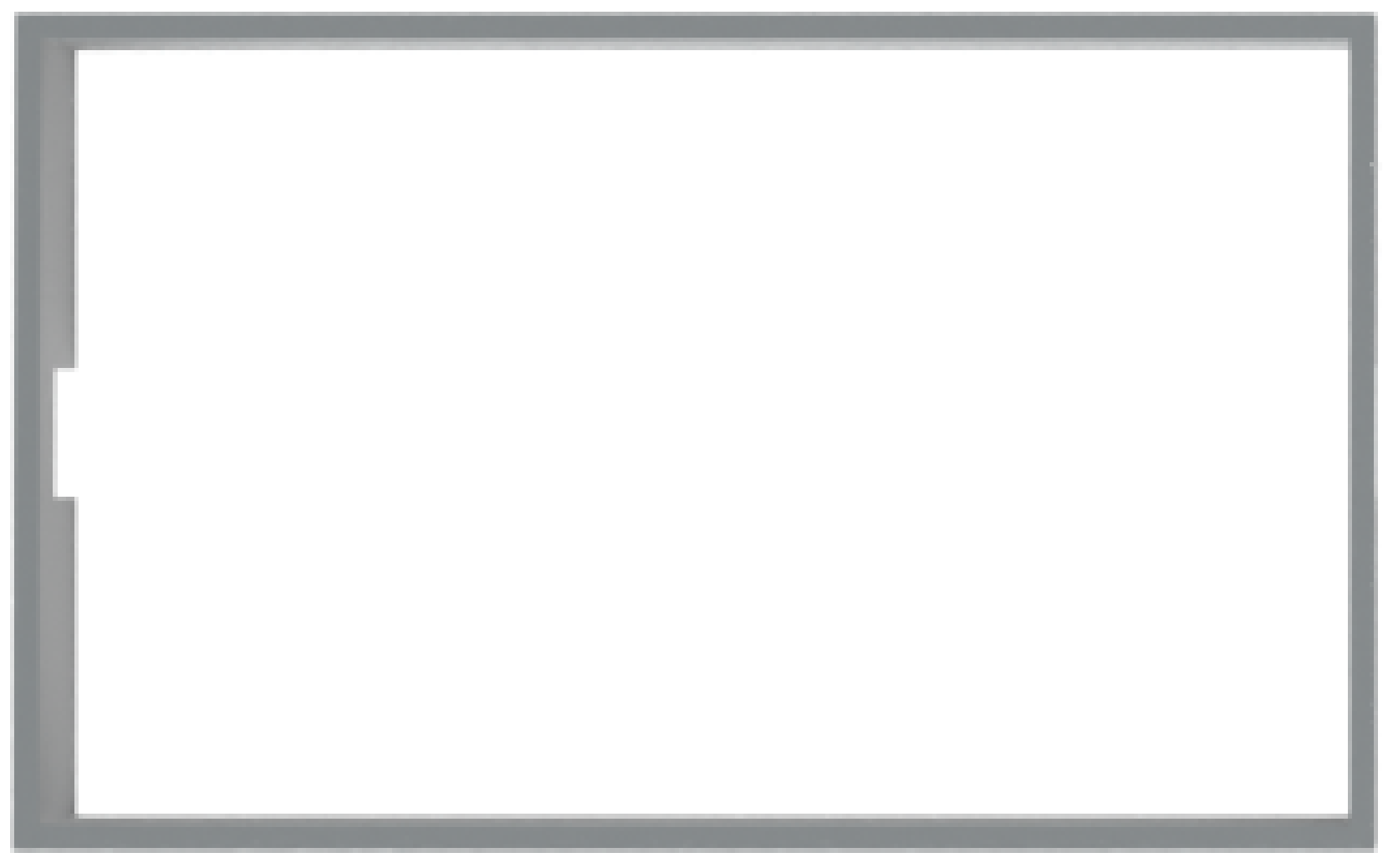}
            %\caption{fig1}
            \label{fig:geo_empty}
        }%
        \subfigure[Structured Space]{
            \centering
            \includegraphics[width=0.32\columnwidth]{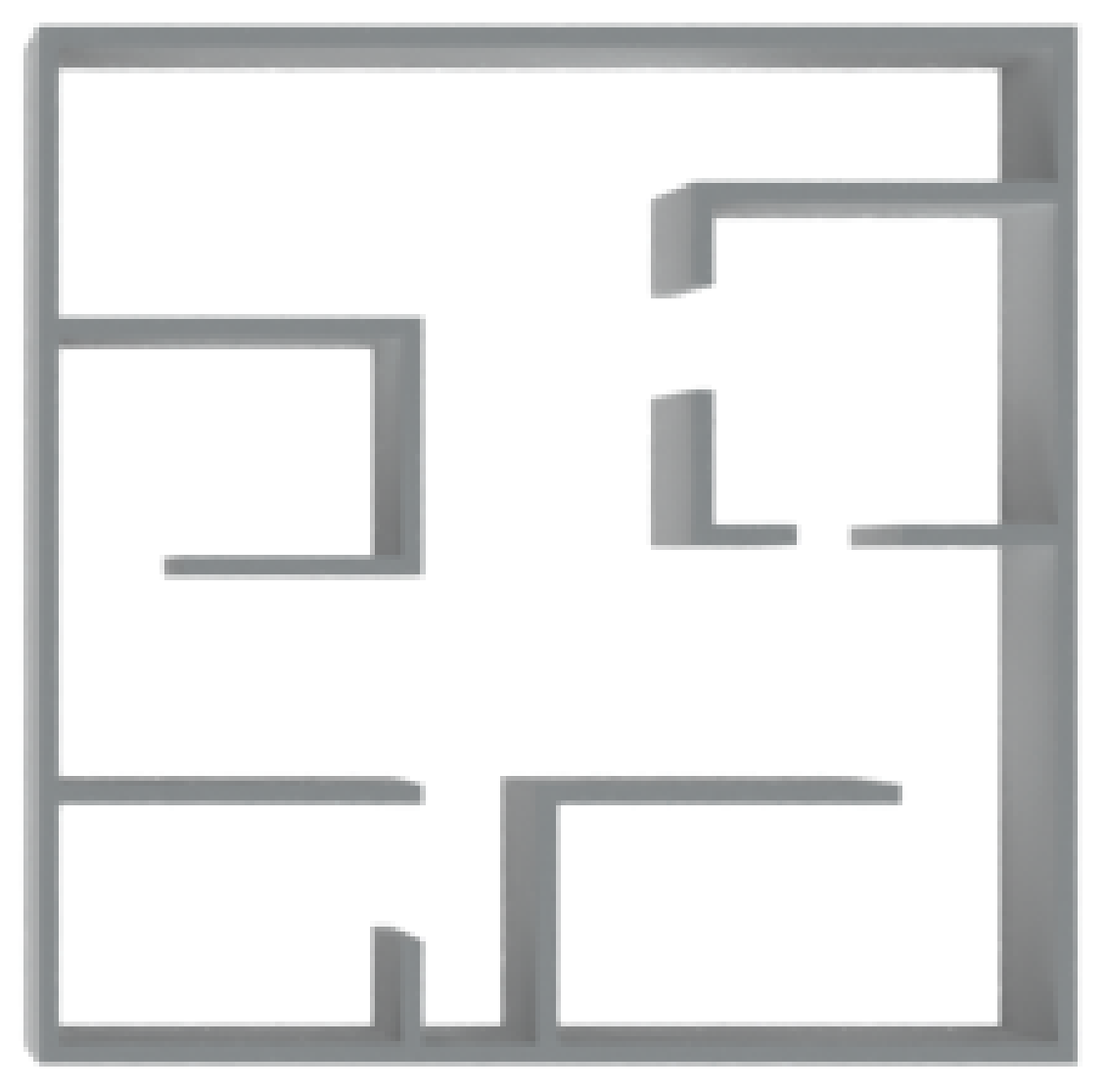}
            \label{fig:geo_room}
        }%
        % \subfigure[Success Rate]{
        %     \centering
        %     \includegraphics[width=0.5\columnwidth]{figs/geometry_successrate.png}
        %     %\caption{fig1}
        %     \label{fig: geo_SR}
        % }%
        % \subfigure[Path Efficiency]{
        %     \centering
        %     % \includegraphics[width=0.5\columnwidth]{figs/geometry_pathefficiency_cdf.png}
        %     \includegraphics[width=0.5\columnwidth]{figs/geometry_pathefficiency_cdf.jpg}
        %     \label{fig: geo_E}
        % }%
    \caption{Illustration of environment geometries. }
    \label{fig:geo_illustration}
\end{figure}

% \begin{figure}
% \setlength{\abovecaptionskip}{0.cm} % height above Figure X caption
% \setlength{\belowcaptionskip}{-0.4cm}
%   \begin{minipage}[t]{0.49\columnwidth}
%     \centering
%     \includegraphics[width=0.98\columnwidth]{figs/D0_searchtime_pathefficiency.png}
%     \caption{Impact of initial distance to source.}
%     \label{fig:ipt_dis}
%   \end{minipage}
%   \begin{minipage}[t]{0.49\columnwidth}
%     \centering
%     \includegraphics[width=0.98\columnwidth]{figs/tau_2robot_ho_successrate_pathefficiency.png}
%     \caption{Impact of temperature parameter.}
%     \label{fig:ipt_tau}
%   \end{minipage}\hfill
% \end{figure}

% \begin{figure}
% \setlength{\abovecaptionskip}{0.cm} % height above Figure X caption
% \setlength{\belowcaptionskip}{-0.4cm}
%   \begin{minipage}[t]{0.49\columnwidth}
%     \centering
%     \includegraphics[width=0.98\columnwidth]{figs/tau_2robot_wVSo_LD.png}
%     \caption{Effectiveness of sensing strategy}
%     \label{fig:eff_sen}
%   \end{minipage}\hfill
%   \begin{minipage}[t]{0.49\columnwidth}
%     \centering
%     \includegraphics[width=0.98\columnwidth]{figs/tau_2robot_hoVShe_successrate_pathefficiency.png}
%     \caption{Effectiveness of planning strategy}
%     \label{fig:eff_plan}
%   \end{minipage}
% \end{figure}

\subsubsection{Impact of Geometry of the Environment}
We evaluate the robustness of our approach in different environment geometries. As shown in \cref{fig:geo_empty,fig:geo_room}, we evaluate all the methods in an empty space (without obstacles) and a structured space (with walls and rooms) to verify the benefits of our method. 
% 1
From \cref{fig: geo}, we can see that \textit{SniffySquad} keeps outperforming the baseline method on both the success rate and path efficiency. 
% 3
Besides, although the structured space significantly increases the difficulty of GSL, our method maintains a success rate of more than $70\%$ and an average path efficiency of around $0.6\%$. Note that Infotaxis shows a better performance than Surge-Cast in the empty space, but its performance degrades greatly when deployed in a more complex structured space. This indicates that \textit{SniffySquad} avoids the oversimplified assumption of either the gas model or the environment, thus featuring the capability to handle complex environments with little performance degradation.

\begin{figure}[t]
    \setlength{\abovecaptionskip}{0.3cm}
    \setlength{\belowcaptionskip}{-0.0cm}
    \centering
    \includegraphics[width=0.8\columnwidth]{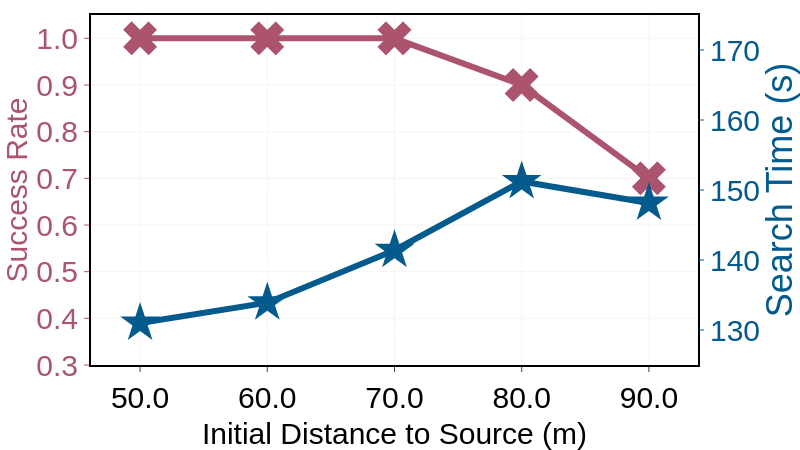}
    \caption{Impact of initial distance to source.}
    \label{fig:ipt_dis}
\end{figure}
\begin{figure}[t]
    \setlength{\abovecaptionskip}{0.3cm}
    \setlength{\belowcaptionskip}{-0.0cm}
    \centering
    \includegraphics[width=0.8\columnwidth]{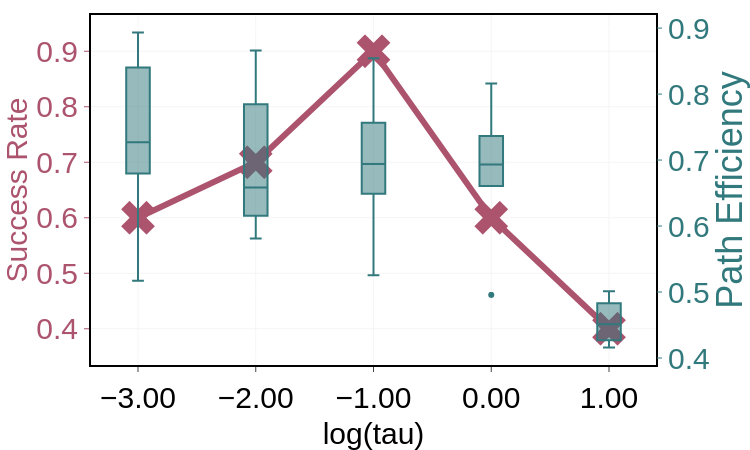}
    \caption{Impact of temperature parameter.}
    \label{fig:ipt_tau}
\end{figure}

\subsubsection{Impact of Initial Distance to the Source}
We examine the impact of robots' initial distances to the gas source on the task. The robots initiate their journey from five distinct positions, ranging from close to gradually increasing distances relative to the gas source. As shown in \cref{fig:ipt_dis}, when the initial distance is less than $70m$, the success rate remains around $100\%$ and the path efficiency decreases gradually as the initial distance grows. However, when it comes to a distance larger than $70m$, we notice that the success rate quickly decreases. This is because the gas plume becomes more and more patchy as the distance increases (as illustrated in \cref{fig:conc_distr_characteristic}), inducing higher probability that the robots are entrapped or misled to a false source localization. We also note that even when the robots start searching at $90m$ away from the source, \textit{SniffySquad} can still achieve an average search time of $148.1s$ for the successful trials.

% \begin{figure*}
% \setlength{\abovecaptionskip}{0.cm} % height above Figure X caption
% \setlength{\belowcaptionskip}{-0.4cm}
%   \begin{minipage}[t]{1.0\columnwidth}
%     \centering
%     \includegraphics[width=0.8\columnwidth]{figs/D0_searchtime_pathefficiency.png}
%     \caption{Impact of initial distance to source.}
%     \label{fig:ipt_dis}
%   \end{minipage}
%   \begin{minipage}[t]{1.0\columnwidth}
%     \centering
%     \includegraphics[width=0.8\columnwidth]{figs/tau_2robot_ho_successrate_pathefficiency.png}
%     \caption{Impact of temperature parameter.}
%     \label{fig:ipt_tau}
%   \end{minipage}

% \end{figure*}

\subsubsection{Impact of Temperature Parameters}
As clarified in \cref{sec:algo2}, the temperature $\tau$ is a critical parameter in the system as it governs the role of a robot. To clarify the impact on the system performance with respect to values of temperature, all the robots are assigned an identical temperature across five orders of magnitude in the experiments.
As illustrated in \cref{fig:ipt_tau}, the results of success rate and path efficiency exhibit different trends as the temperature rises. For the success rate, as the temperature increases, it shows an incline trend at first and reaches the maximum at the temperature of $10^{-1}$, and then declines gradually; for the path efficiency, its value gradually decreases as the temperature grows, especially when it rises over $10^{-1}$. This pattern echoes our previous analysis that either too exploitative movement (low temperature) or explorative movement (high temperature) fails to balance the trade-off of search efficiency and effectiveness. \textit{SniffySquad} combines the advantages of both behaviors via the collaborative role-switching method, making \textit{SniffySquad} more superior to other approaches.

%% file: conclusion.tex
\section{Conclusion} \label{sec:conclusion}
This paper proposes \textit{SniffySquad}, a multi-robot olfaction-based system for gas source localization. 
\textit{SniffySquad} features two key designs from non-convex learning perspective to conquer the patchy characteristic of gas plumes and enable efficient and effective source localization by a collaborative strategy, as elaborated below. 
Specifically, we devise a patchy plume-resilient movement strategy inspired by Langevin diffusion approach. 
We then develop a collaborative strategy for multiple robots by assigning and adapting their planning preferences dynamically to balance the search efficiency and localization effectiveness of gas sources.
We implement \textit{SniffySquad} with multiple unmanned ground vehicles and conduct experiments in a real-world testbed. Results show that \textit{SniffySquad} achieves a remarkable $20\%+$ success rate and $30\%+$ path efficiency improvement, respectively, outperforming state-of-the-art gas source localization solutions.

%% file: main.bbl
\begin{thebibliography}{10}
\providecommand{\url}[1]{#1}
\csname url@samestyle\endcsname
\providecommand{\newblock}{\relax}
\providecommand{\bibinfo}[2]{#2}
\providecommand{\BIBentrySTDinterwordspacing}{\spaceskip=0pt\relax}
\providecommand{\BIBentryALTinterwordstretchfactor}{4}
\providecommand{\BIBentryALTinterwordspacing}{\spaceskip=\fontdimen2\font plus
\BIBentryALTinterwordstretchfactor\fontdimen3\font minus \fontdimen4\font\relax}
\providecommand{\BIBforeignlanguage}[2]{{%
\expandafter\ifx\csname l@#1\endcsname\relax
\typeout{** WARNING: IEEEtran.bst: No hyphenation pattern has been}%
\typeout{** loaded for the language `#1'. Using the pattern for}%
\typeout{** the default language instead.}%
\else
\language=\csname l@#1\endcsname
\fi
#2}}
\providecommand{\BIBdecl}{\relax}
\BIBdecl

\bibitem{liu2017individualized}
X.~Liu, X.~Xu, X.~Chen, E.~Mai, H.~Y. Noh, P.~Zhang, and L.~Zhang, ``Individualized calibration of industrial-grade gas sensors in air quality sensing system,'' in \emph{Proceedings of the 15th ACM Conference on Embedded Network Sensor Systems}, 2017, pp. 1--2.

\bibitem{pal2018vehicle}
S.~Pal, A.~Ghosh, and V.~Sethi, ``Vehicle air pollution monitoring using iots,'' in \emph{Proceedings of the 16th ACM conference on embedded networked sensor systems}, 2018, pp. 400--401.

\bibitem{maag2018w}
B.~Maag, Z.~Zhou, and L.~Thiele, ``W-air: Enabling personal air pollution monitoring on wearables,'' \emph{Proceedings of the ACM on Interactive, Mobile, Wearable and Ubiquitous Technologies}, vol.~2, no.~1, pp. 1--25, 2018.

\bibitem{sun2024gastag}
X.~Sun, J.~Xiong, C.~Feng, X.~Li, J.~Zhang, B.~Li, D.~Fang, and X.~Chen, ``Gastag: A gas sensing paradigm using graphene-based tags,'' in \emph{Proceedings of the 30th Annual International Conference on Mobile Computing and Networking}, 2024, pp. 342--356.

\bibitem{liu2017delay}
X.~Liu, X.~Chen, X.~Xu, E.~Mai, H.~Y. Noh, P.~Zhang, and L.~Zhang, ``Delay effect in mobile sensing system for urban air pollution monitoring,'' in \emph{Proceedings of the 15th ACM Conference on Embedded Network Sensor Systems}, 2017, pp. 1--2.

\bibitem{mccarron2018air}
G.~McCarron, ``Air pollution and human health hazards: a compilation of air toxins acknowledged by the gas industry in queensland’s darling downs,'' \emph{International Journal of Environmental Studies}, vol.~75, no.~1, pp. 171--185, 2018.

\bibitem{manes2016realtime}
G.~Manes, G.~Collodi, L.~Gelpi, R.~Fusco, G.~Ricci, A.~Manes, and M.~Passafiume, ``Realtime gas emission monitoring at hazardous sites using a distributed point-source sensing infrastructure,'' \emph{Sensors}, vol.~16, no.~1, p. 121, 2016.

\bibitem{bourne2020decentralized}
J.~R. Bourne, M.~N. Goodell, X.~He, J.~A. Steiner, and K.~K. Leang, ``Decentralized multi-agent information-theoretic control for target estimation and localization: finding gas leaks,'' \emph{The International Journal of Robotics Research}, vol.~39, no.~13, pp. 1525--1548, 2020.

\bibitem{hutchinson2018entrotaxis}
M.~Hutchinson, H.~Oh, and W.-H. Chen, ``Entrotaxis as a strategy for autonomous search and source reconstruction in turbulent conditions,'' \emph{Information Fusion}, vol.~42, pp. 179--189, 2018.

\bibitem{reddy2022olfactory}
G.~Reddy, V.~N. Murthy, and M.~Vergassola, ``Olfactory sensing and navigation in turbulent environments,'' \emph{Annual Review of Condensed Matter Physics}, vol.~13, pp. 191--213, 2022.

\bibitem{chen2019odor}
X.-x. Chen and J.~Huang, ``Odor source localization algorithms on mobile robots: A review and future outlook,'' \emph{Robotics and Autonomous Systems}, vol. 112, pp. 123--136, 2019.

\bibitem{rhodes2023autonomous}
C.~Rhodes, C.~Liu, P.~Westoby, and W.-H. Chen, ``Autonomous search of an airborne release in urban environments using informed tree planning,'' \emph{Autonomous Robots}, vol.~47, no.~1, pp. 1--18, 2023.

\bibitem{rhodes2020informative}
C.~Rhodes, C.~Liu, and W.-H. Chen, ``Informative path planning for gas distribution mapping in cluttered environments,'' in \emph{2020 IEEE/RSJ International Conference on Intelligent Robots and Systems (IROS)}.\hskip 1em plus 0.5em minus 0.4em\relax IEEE, 2020, pp. 6726--6732.

\bibitem{bourne2019CoordinatedBayesianBasedBioinspired}
J.~R. Bourne, E.~R. Pardyjak, and K.~K. Leang, ``Coordinated {{Bayesian-Based Bioinspired Plume Source Term Estimation}} and {{Source Seeking}} for {{Mobile Robots}},'' \emph{IEEE Transactions on Robotics}, vol.~35, no.~4, pp. 967--986, 2019.

\bibitem{duisterhof2021sniffy}
B.~P. Duisterhof, S.~Li, J.~Burgu{\'e}s, V.~J. Reddi, and G.~C. de~Croon, ``Sniffy bug: A fully autonomous swarm of gas-seeking nano quadcopters in cluttered environments,'' in \emph{2021 IEEE/RSJ International Conference on Intelligent Robots and Systems (IROS)}.\hskip 1em plus 0.5em minus 0.4em\relax IEEE, 2021, pp. 9099--9106.

\bibitem{deng2020contourSGLD}
W.~Deng, G.~Lin, and F.~Liang, ``A contour stochastic gradient langevin dynamics algorithm for simulations of multi-modal distributions,'' \emph{Advances in neural information processing systems}, vol.~33, pp. 15\,725--15\,736, 2020.

\bibitem{loisy2022searching}
A.~Loisy and C.~Eloy, ``Searching for a source without gradients: how good is infotaxis and how to beat it,'' \emph{Proceedings of the Royal Society A}, vol. 478, no. 2262, p. 20220118, 2022.

\bibitem{holzbecher20122d}
E.~Holzbecher and E.~Holzbecher, ``2d and 3d transport solutions (gaussian puffs and plumes),'' \emph{Environmental modeling: using MATLAB}, pp. 303--316, 2012.

\bibitem{vergassola2007infotaxis}
M.~Vergassola, E.~Villermaux, and B.~I. Shraiman, ``‘infotaxis’ as a strategy for searching without gradients,'' \emph{Nature}, vol. 445, no. 7126, pp. 406--409, 2007.

\bibitem{wang2018efficient}
C.~Wang, T.~Li, M.~Q.-H. Meng, and C.~De~Silva, ``Efficient mobile robot exploration with gaussian markov random fields in 3d environments,'' in \emph{2018 IEEE International Conference on Robotics and Automation (ICRA)}.\hskip 1em plus 0.5em minus 0.4em\relax IEEE, 2018, pp. 5015--5021.

\bibitem{lilienthal2009statistical}
A.~J. Lilienthal, M.~Reggente, M.~Trincavelli, J.~L. Blanco, and J.~Gonzalez, ``A statistical approach to gas distribution modelling with mobile robots-the kernel dm+ v algorithm,'' in \emph{2009 IEEE/RSJ International Conference on Intelligent Robots and Systems}.\hskip 1em plus 0.5em minus 0.4em\relax IEEE, 2009, pp. 570--576.

\bibitem{gongora2020joint}
A.~Gongora, J.~Monroy, and J.~Gonzalez-Jimenez, ``Joint estimation of gas and wind maps for fast-response applications,'' \emph{Applied Mathematical Modelling}, vol.~87, pp. 655--674, 2020.

\bibitem{bourne2019coordinated}
J.~R. Bourne, E.~R. Pardyjak, and K.~K. Leang, ``Coordinated bayesian-based bioinspired plume source term estimation and source seeking for mobile robots,'' \emph{IEEE Transactions on Robotics}, vol.~35, no.~4, pp. 967--986, 2019.

\bibitem{marques2002olfaction}
L.~Marques, U.~Nunes, and A.~T. de~Almeida, ``Olfaction-based mobile robot navigation,'' \emph{Thin solid films}, vol. 418, no.~1, pp. 51--58, 2002.

\bibitem{cao2023catnipp}
Y.~Cao, Y.~Wang, A.~Vashisth, H.~Fan, and G.~A. Sartoretti, ``Catnipp: Context-aware attention-based network for informative path planning,'' in \emph{Conference on Robot Learning}.\hskip 1em plus 0.5em minus 0.4em\relax PMLR, 2023, pp. 1928--1937.

\bibitem{Ma2023GaussianME}
\BIBentryALTinterwordspacing
H.~Ma, T.~Zhang, Y.~Wu, F.~du~Pin~Calmon, and N.~Li, ``Gaussian max-value entropy search for multi-agent bayesian optimization,'' \emph{2023 IEEE/RSJ International Conference on Intelligent Robots and Systems (IROS)}, pp. 10\,028--10\,035, 2023. [Online]. Available: \url{https://api.semanticscholar.org/CorpusID:257482432}
\BIBentrySTDinterwordspacing

\bibitem{Zhang2022DistributedIS}
\BIBentryALTinterwordspacing
T.~Zhang, V.~Qin, Y.~Tang, and N.~Li, ``Distributed information-based source seeking,'' \emph{IEEE Transactions on Robotics}, vol.~39, pp. 4749--4767, 2022. [Online]. Available: \url{https://api.semanticscholar.org/CorpusID:252383539}
\BIBentrySTDinterwordspacing

\bibitem{kingma2014adam}
D.~P. Kingma and J.~Ba, ``Adam: A method for stochastic optimization,'' \emph{arXiv preprint arXiv:1412.6980}, 2014.

\bibitem{cheng2020interplay}
X.~Cheng, \emph{The Interplay between Sampling and Optimization}.\hskip 1em plus 0.5em minus 0.4em\relax University of California, Berkeley, 2020.

\bibitem{ma2019samplingcanbefaster}
Y.-A. Ma, Y.~Chen, C.~Jin, N.~Flammarion, and M.~I. Jordan, ``Sampling can be faster than optimization,'' \emph{Proceedings of the National Academy of Sciences}, vol. 116, no.~42, pp. 20\,881--20\,885, 2019.

\bibitem{chen2020accelerating}
Y.~Chen, J.~Chen, J.~Dong, J.~Peng, and Z.~Wang, ``Accelerating nonconvex learning via replica exchange langevin diffusion,'' \emph{arXiv preprint arXiv:2007.01990}, 2020.

\bibitem{strikwerda2004finite}
J.~C. Strikwerda, \emph{Finite difference schemes and partial differential equations}.\hskip 1em plus 0.5em minus 0.4em\relax SIAM, 2004.

\bibitem{ojeda2021information}
P.~Ojeda, J.~Monroy, and J.~Gonzalez-Jimenez, ``Information-driven gas source localization exploiting gas and wind local measurements for autonomous mobile robots,'' \emph{IEEE Robotics and Automation Letters}, vol.~6, no.~2, pp. 1320--1326, 2021.

\bibitem{earl2005paralleltempering}
D.~J. Earl and M.~W. Deem, ``Parallel tempering: Theory, applications, and new perspectives,'' \emph{Physical Chemistry Chemical Physics}, vol.~7, no.~23, pp. 3910--3916, 2005.

\bibitem{hutchinson2019experimental}
M.~Hutchinson, P.~Ladosz, C.~Liu, and W.-H. Chen, ``Experimental assessment of plume mapping using point measurements from unmanned vehicles,'' in \emph{2019 International Conference on Robotics and Automation (ICRA)}.\hskip 1em plus 0.5em minus 0.4em\relax IEEE, 2019, pp. 7720--7726.

\bibitem{hutchinson2018information}
M.~Hutchinson, C.~Liu, and W.-H. Chen, ``Information-based search for an atmospheric release using a mobile robot: Algorithm and experiments,'' \emph{IEEE Transactions on Control Systems Technology}, vol.~27, no.~6, pp. 2388--2402, 2018.

\bibitem{kadakia2022odour}
N.~Kadakia, M.~Demir, B.~T. Michaelis, B.~D. DeAngelis, M.~A. Reidenbach, D.~A. Clark, and T.~Emonet, ``Odour motion sensing enhances navigation of complex plumes,'' \emph{Nature}, vol. 611, no. 7937, pp. 754--761, 2022.

\bibitem{francis2022gas}
A.~Francis, S.~Li, C.~Griffiths, and J.~Sienz, ``Gas source localization and mapping with mobile robots: A review,'' \emph{Journal of Field Robotics}, vol.~39, no.~8, pp. 1341--1373, 2022.

\bibitem{levy2018field}
M.~Levy~Zamora, F.~Xiong, D.~Gentner, B.~Kerkez, J.~Kohrman-Glaser, and K.~Koehler, ``Field and laboratory evaluations of the low-cost plantower particulate matter sensor,'' \emph{Environmental science \& technology}, vol.~53, no.~2, pp. 838--849, 2018.

\bibitem{jasak2009openfoam}
H.~Jasak, ``Openfoam: Open source cfd in research and industry,'' \emph{International Journal of Naval Architecture and Ocean Engineering}, vol.~1, no.~2, pp. 89--94, 2009.

\bibitem{monroy2017gaden}
J.~Monroy, V.~Hernandez-Bennetts, H.~Fan, A.~Lilienthal, and J.~Gonzalez-Jimenez, ``Gaden: A 3d gas dispersion simulator for mobile robot olfaction in realistic environments,'' \emph{Sensors}, vol.~17, no.~7, p. 1479, 2017.

\bibitem{farrell2002filament}
J.~A. Farrell, J.~Murlis, X.~Long, W.~Li, and R.~T. Card{\'e}, ``Filament-based atmospheric dispersion model to achieve short time-scale structure of odor plumes,'' \emph{Environmental fluid mechanics}, vol.~2, pp. 143--169, 2002.

\bibitem{chen2024adaptive}
W.~Chen, R.~Khardon, and L.~Liu, ``Adaptive robotic information gathering via non-stationary gaussian processes,'' \emph{The International Journal of Robotics Research}, vol.~43, no.~4, pp. 405--436, 2024.

\end{thebibliography}
